\documentclass[11pt]{article}

\usepackage[preprint]{acl}

\usepackage{times}
\usepackage{latexsym}

\usepackage[T1]{fontenc}

\usepackage[utf8]{inputenc}

\usepackage{microtype}
\usepackage{inconsolata}
\usepackage{graphicx}

\usepackage{booktabs}
\usepackage{tabularx}
\usepackage{multirow}
\usepackage{enumitem}
\usepackage[most]{tcolorbox}
\usepackage{xcolor}

\newtcolorbox{promptbox}{
  enhanced,
  breakable,
  colback=gray!5,
  colframe=gray!50,
  boxrule=0.5pt,
  arc=2pt,
  left=4pt, right=4pt, top=4pt, bottom=4pt,
  before skip=6pt,
  after skip=6pt,
  before upper={\parindent0pt\raggedright\sloppy},
}
\newcommand{\promptheading}[1]{\vspace{0.5em}\noindent\textbf{#1}\par\vspace{0.25em}}

\newcommand{\best}[1]{\textbf{#1}}

\title{AuthTrace: Diagnosing Evidence Construction in Thematically Dense Single-Author Corpora}

\author{
Xiaoqing Wu \and Feifei Li \and Haoliang Ming \and Wenhui Que \\
WeChat, Tencent Inc., Beijing, China \\
\texttt{\{xiaoqingwwu, niyali, hliangming, victorque\}@tencent.com}
}

\begin{document}
\maketitle

\begin{abstract}
Evidence construction—the stage that determines which passages reach the language model before generation begins—is evaluated paradigm by paradigm, leaving practitioners with no principled way to diagnose which organization strategy fails, where, or why. We introduce \textbf{AuthTrace}, a diagnostic benchmark built on thematically dense single-author corpora where near-miss distractors share style, topic, and vocabulary with the required evidence. AuthTrace provides explicit quoted evidence, exact fan-in annotation, and a unified pack-level protocol measuring evidence recall, evidence precision, and answer correctness. A fan-in gradient---the number of source documents required to support the answer---serves as the primary diagnostic axis, enabling controlled comparison across retrieval, memory, graph, and structured-evidence paradigms. Evaluating eight systems across two QA models, we find that evidence recall is the strongest observed predictor of answer correctness under the primary reader–judge pair (r = 0.96); most failures stem from missing evidence rather than answer synthesis. Fan-in further exposes paradigm-specific collapse patterns: flat retrieval degrades 2–3× faster than thematically organized evidence construction. These results show fan-in decomposition to be a reusable diagnostic lens for identifying where evidence-construction systems fail and which paradigm best serves a given workload.
\end{abstract}

\section{Introduction}

Evidence construction is increasingly handled by heterogeneous organization strategies—flat retrieval, memory compression, graph traversal, and structured thematic indexing—yet these paradigms are benchmarked in isolation: each brings its own corpus, output unit, and evaluation metric. As a result, when an evidence system fails, practitioners cannot determine whether the failure stems from missing evidence, noisy evidence, or flawed answer synthesis, nor can they identify which organization paradigm would better serve the workload. A shared diagnostic protocol is needed. We observe that a single-author corpus can be instantiated under each of these evidence-organization views: it is a bounded document collection amenable to chunking and embedding \citep{karpukhin-etal-2020-dense,NEURIPS2020_6b493230}, an accumulated knowledge source suitable for memory compression \citep{packer2024memgptllmsoperatingsystems}, a relational structure for graph construction \citep{NEURIPS2024_6ddc001d}, and a thematic archive for wiki organization \citep{ming2026retrievalreasoningselfevolvingagentnative}.

The setting is especially challenging when the corpus is thematically dense. In open-domain retrieval and heterogeneous IR evaluation, irrelevant passages often differ from the query in topic, entities, or genre \citep{thakur2021beir,yang-etal-2018-hotpotqa}. In an author-centered corpus, many distractors are near misses: they share the same style, vocabulary, historical setting, and recurring concerns, yet they do not directly support the target claim. Evidence selection in this setting requires distinguishing direct support from plausible thematic adjacency, and many questions require support to be assembled across several documents.

\begin{figure*}[t]
\centering
\includegraphics[width=\textwidth]{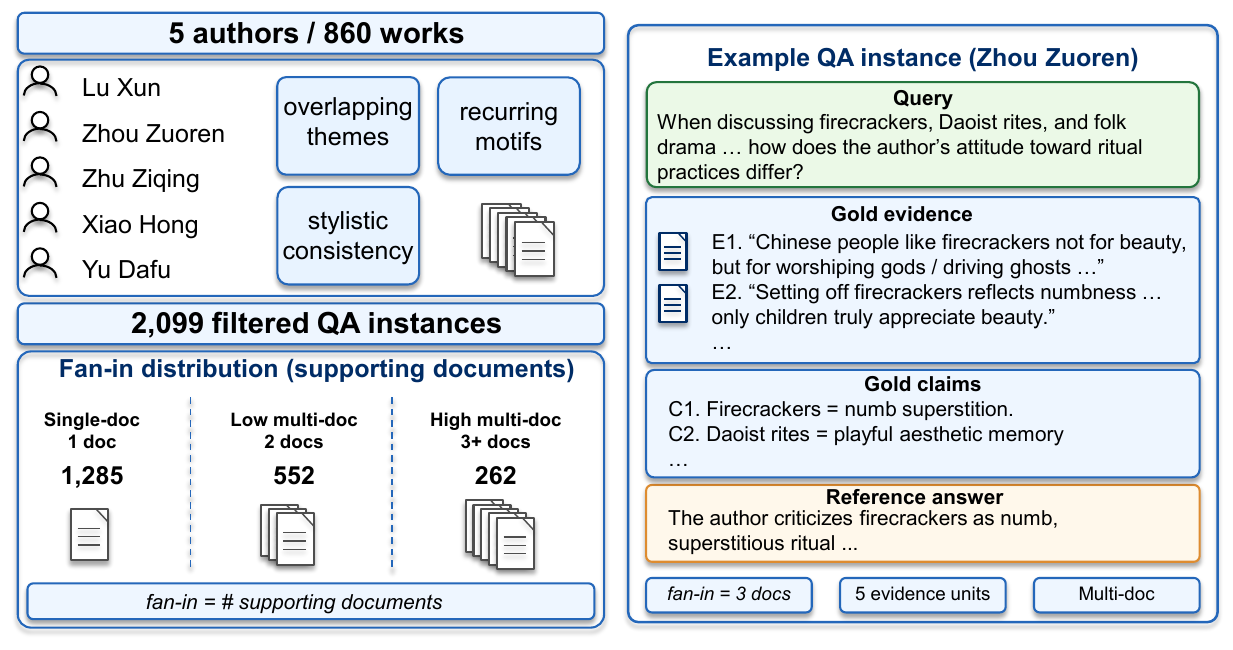}
\caption{AuthTrace benchmark overview. The dataset spans five authors and 860 works, yielding 2,099 filtered QA instances with exact fan-in labels. Each instance contains a query, quoted gold evidence, atomic gold claims, a reference answer, and a fan-in annotation that records the number of supporting documents.}
\label{fig:dataset-overview}
\end{figure*}

\begin{figure*}[t]
\centering
\includegraphics[width=0.75\textwidth,height=0.32\textheight,keepaspectratio]{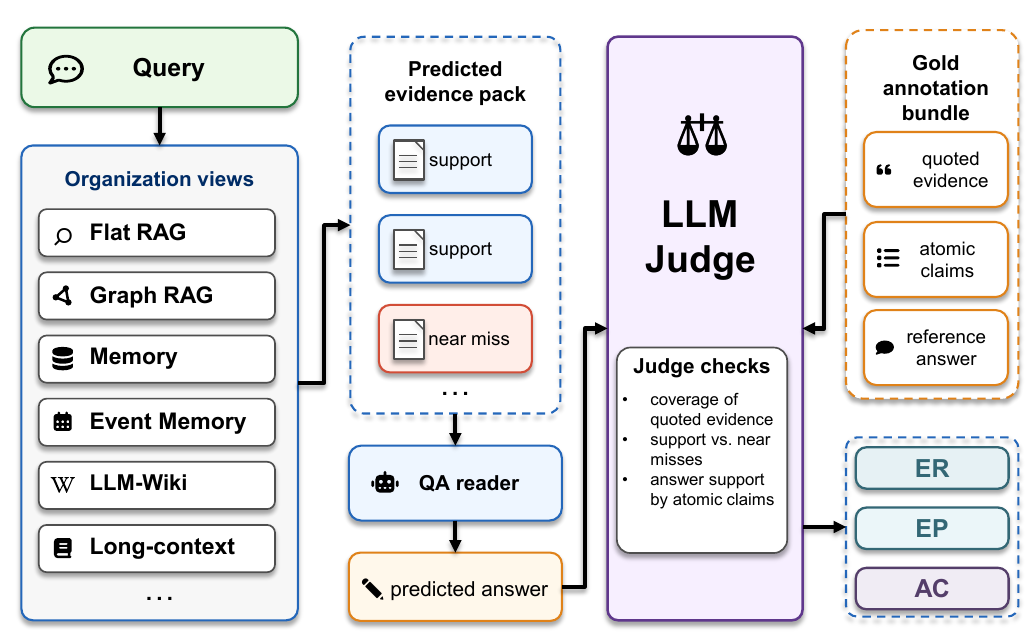}
\caption{AuthTrace evaluation framework. System-specific organization views produce a predicted evidence pack and answer, which are judged against the gold annotation bundle using Evidence Recall (ER), Evidence Precision (EP), and Answer Correctness (AC).}
\label{fig:evaluation-framework}
\end{figure*}

We introduce \textbf{AuthTrace}, a diagnostic benchmark for evidence construction in thematically dense single-author corpora. AuthTrace is built from 860 public-domain writings by five modern Chinese prose writers. Each instance provides a query, quoted gold evidence units linked to source documents, atomic gold claim units, a reference answer, and an exact evidence fan-in label---the number of distinct source documents required to support the answer. Fan-in is the benchmark's main diagnostic axis, grouping instances into Single-doc ($=1$), Low multi-doc ($=2$), and High multi-doc ($\geq 3$). All systems are evaluated through a unified pack-level protocol measuring Evidence Recall (ER), Evidence Precision (EP), and Answer Correctness (AC). Figures~\ref{fig:dataset-overview} and~\ref{fig:evaluation-framework} summarize the benchmark composition and evaluation framework.

Through this design, AuthTrace enables the first controlled cross-paradigm comparison of evidence construction. We make three contributions:
\begin{enumerate}[leftmargin=*]
    \item A unified cross-paradigm diagnostic framework that places retrieval, memory, graph, and structured-evidence systems on the same corpus, queries, and pack-level protocol for controlled comparison across previously siloed evaluation settings.
    \item AuthTrace, a publicly released benchmark across five single-author corpora with quoted required evidence, exact fan-in annotation, and human-validated accuracy---coupled with a fan-in complexity gradient that localizes each system's breaking point.\footnote{Data, corpus, and evaluation code: \url{https://anonymous.4open.science/r/AuthTrace-anonymous-7BD8}}
    \item Empirical findings showing that (i) no paradigm dominates—graph retrieval excels at local grounding while thematic indexing leads on multi-document synthesis, with a crossover at fan-in 2; (ii) evidence recall is the main observed failure mode; and (iii) fan-in exposes paradigm-specific collapse rates, yielding a selection heuristic: graph retrieval for $f=1$, thematic indexing for $f\geq2$.
\end{enumerate}

\section{AuthTrace Task}

Given an author corpus $\mathcal{C}_a = \{d_1, \ldots, d_n\}$ and a query $q$, a system outputs an evidence pack $\hat{E}_q$ and an answer $\hat{a}$. Each benchmark instance provides a reference answer $a^\star$, gold claim units $C^\star$, and gold evidence units $E^\star$. Each evidence unit is a quoted span paired with a source document, making the required support explicit at document and passage level.

We define evidence fan-in as the number of distinct source documents appearing in the gold evidence:
\begin{equation}
    \mathrm{fan\mbox{-}in}(q) = \left| \{ d : (d, e) \in E^\star \} \right| .
\end{equation}
We group examples into \textbf{Single-doc} ($=1$), \textbf{Low multi-doc} ($=2$), and \textbf{High multi-doc} ($\geq 3$). Single-doc questions test local grounding, low-fan-in questions test small-scale aggregation, and high-fan-in questions test broad synthesis. Fan-in is not intended to exhaustively parameterize question difficulty; it isolates an interpretable dimension of evidence-construction complexity.

The evidence pack is evaluated as a unit. A strong pack must cover the evidence required by the answer while avoiding passages that are merely thematically related. This criterion is stricter than ordinary passage relevance or citation evaluation: a passage may discuss the same theme and still be invalid evidence for the specific claim being asked \citep{gao-etal-2023-enabling,es-etal-2024-ragas}. This separation lets AuthTrace distinguish failures in evidence selection, organization, and answer synthesis. Additional motivation for the single-author setting and the cross-paradigm interpretation appears in Appendix~\ref{app:task-details}.

\section{Metrics}

\paragraph{Answer Correctness.}
Answer Correctness (AC) uses a holistic 0--3 rubric (0 = incorrect or irrelevant, 3 = complete and precise; full rubric in Appendix~\ref{app:ac-prompt}) that jointly penalizes missing gold claims and irrelevant content. The LLM judge receives the query, reference answer, and gold claim units as a scoring guide, then rates the predicted answer on claim coverage and factual precision. The reported AC metric is normalized to $[0, 100]$: $\text{AC} = \text{score} / 3 \times 100$.

\paragraph{Evidence Recall.}
Evidence Recall (ER) measures how much of the gold evidence $E^\star$ is covered by the predicted evidence pack $\hat{E}_q$:
\begin{equation}
    \mathrm{ER} = \frac{|\mathrm{matched}(E^\star, \hat{E}_q)|}{|E^\star|}.
\end{equation}
A gold unit is counted as matched when an LLM judge confirms that the predicted pack contains a passage covering its key factual content (binary decision; prompt in Appendix~\ref{app:er-prompt}). ER diagnoses missing-evidence failures: low ER indicates that the answer model did not receive enough support, regardless of how focused the retrieved pack is.

\paragraph{Evidence Precision.}
Evidence Precision (EP) measures the fraction of predicted evidence units in $\hat{E}_q$ that directly support at least one gold evidence unit or its associated atomic claim:
\begin{equation}
    \mathrm{EP} = \frac{|\mathrm{supporting}(\hat{E}_q, E^\star)|}{|\hat{E}_q|}.
\end{equation}
A predicted unit is counted as supporting when the judge confirms it can be aligned with at least one gold entry (prompt in Appendix~\ref{app:ep-prompt}). EP diagnoses topical over-selection and context inefficiency. High ER with low EP indicates over-retrieval with sufficient support; high EP with low ER indicates a clean but incomplete pack. Appendix~\ref{app:metric-interpretation} discusses the context-budget interpretation of EP.

\section{Dataset}

AuthTrace emphasizes annotation depth. Each accepted instance contains quoted required evidence, document-level provenance, atomic claims, a concise reference answer, and exact fan-in. This design supports direct computation of pack-level ER and EP, which cannot be recovered from answer-only labels or loose relevance annotations \citep{gao-etal-2023-enabling,saad-falcon-etal-2024-ares}.

AuthTrace is built from public-domain writings by Lu Xun, Zhou Zuoren, Zhu Ziqing, Xiao Hong, and Yu Dafu---860 articles totaling approximately 2M Chinese characters. These writers provide coherent but thematically dense corpora: each returns across essays to overlapping social, cultural, literary, and personal concerns (per-author statistics in Appendix~\ref{app:dataset-pipeline}, Table~\ref{tab:corpus}).

AuthTrace is constructed through a staged LLM-assisted annotation pipeline. We first curate the author corpora at the document level, retaining nonfiction prose in which the author's own viewpoint occupies the dominant share of the text and removing documents shorter than 800 Chinese characters. Each retained article is assigned a primary theme from a fixed taxonomy of 11 canonical themes. Single-document instances are generated from individual high-quality articles, while multi-document instances are generated from same-theme document groups sampled to create low- and high-fan-in evidence requirements. Appendix~\ref{app:dataset-pipeline} gives the full generation and normalization protocol.

Each instance stores a query with title leakage removed, quoted gold evidence units, atomic gold claim units, a concise reference answer, and exact fan-in computed from the distinct documents appearing in the gold evidence (full annotation schema in Appendix~\ref{app:annotation}; representative examples in Appendix~\ref{app:examples}). The final benchmark is obtained by filtering generated instances for query validity, evidence-answer support, and quote--document consistency (generation constraints and prompts in Appendix~\ref{app:prompts} and \ref{app:generation-prompts}). After filtering, the benchmark contains 2,099 instances: 1,285 single-doc ($f{=}1$), 552 low multi-doc ($f{=}2$), and 262 high multi-doc ($f{\geq}3$).

A human spot-check on 200 randomly sampled instances finds 97.0\% overall validity (100\% query validity, 97.0\% evidence-answer support, 100\% quote--document consistency). The remaining invalid cases are primarily evidence-support errors, where the quoted evidence is related to the reference answer but does not fully support all required claims.

\section{Experimental Setup}

We compare evidence construction across eight systems and baselines: closed-book answering, long-context prompting, oracle evidence, flat RAG \citep{NEURIPS2020_6b493230}, HippoRAG2 \citep{pmlr-v267-gutierrez25a}, Mem0 \citep{DBLP:conf/ecai/ChhikaraKASY25}, EverMemOS \citep{hu2026evermemosselforganizingmemoryoperating}, and LLM-Wiki evidence construction \citep{ming2026retrievalreasoningselfevolvingagentnative}. Each paradigm imposes a different corpus view and evidence-construction mechanism (summarized in Appendix~\ref{app:task-details}, Table~\ref{tab:paradigms}). All systems operate over the same cleaned author corpora and answer the same question set. Retrieval-based and organization-based systems return an evidence pack that is evaluated for ER and EP before answer generation. Closed-book systems do not output evidence packs and are evaluated only by AC. Long-context prompting is treated as a separate baseline because it tests large-budget sequential corpus exposure rather than compact evidence construction \citep{bai-etal-2024-longbench,hsieh2024ruler,zhang-etal-2024-bench}.

The main experiment uses GLM-5.1-FP8 \citep{glm5team2026glm5vibecodingagentic} as the QA reader and Qwen3.5-397B-A17B \citep{qwen35blog} as the evaluator, separating answer generation from judgment. We further run a model-family ablation on a 20\% stratified sample (419 instances) that crosses two QA readers (GLM-5.1-FP8, GPT-4.1-mini) with two judges (Qwen3.5-397B-A17B, GPT-4o-mini), holding evidence packs fixed to isolate reader and judge effects. For systems that produce ranked evidence units, the evidence budget is held fixed where possible, and evidence units are normalized into the same matching protocol, following the broader practice of comparing retrieval systems under shared evaluation interfaces \citep{thakur2021beir,petroni-etal-2021-kilt}. Full system descriptions and implementation details appear in Appendix~\ref{app:implementation}.

\paragraph{Evidence matching.}
Predicted evidence packs are normalized into canonical segments before evaluation (normalization details in Appendix~\ref{app:evaluation-protocol}). The matching protocol treats near-miss thematic passages as non-supporting even when they are topically close, ensuring that ER and EP reflect genuine evidence coverage rather than mere topical adjacency. Human--LLM agreement validation (Appendix~\ref{sec:evaluation-reliability}) confirms the reliability of the LLM judge: Spearman $\rho = 0.78$ for AC, 91.2\% binary agreement for ER, and 88.4\% for EP.

\section{Results}

\subsection{Evidence Construction Is Recall-Dominant in the Main Setting}

\begin{figure}[t]
\centering
\includegraphics[width=0.8\columnwidth]{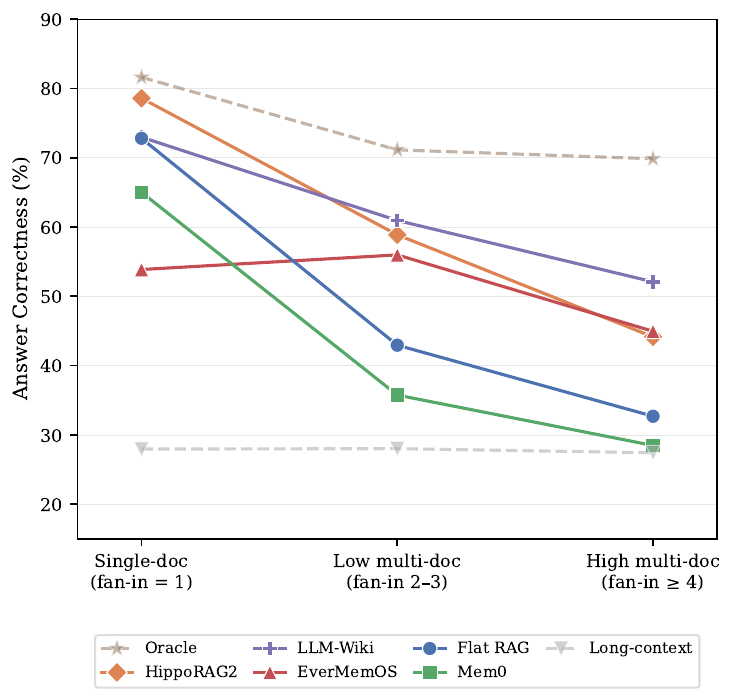}
\caption{Fan-in degradation curves for evidence-construction paradigms. Each curve traces AC from Single-doc through Low multi-doc to High multi-doc, revealing characteristic degradation profiles: cliff-like (Flat RAG, Mem0), stepwise (HippoRAG2), gradual (LLM-Wiki), and rebound (EverMemOS, which increases at Low multi-doc before declining).}
\label{fig:fanin-degradation}
\end{figure}

\begin{figure}[t]
\centering
\includegraphics[width=1.0\columnwidth]{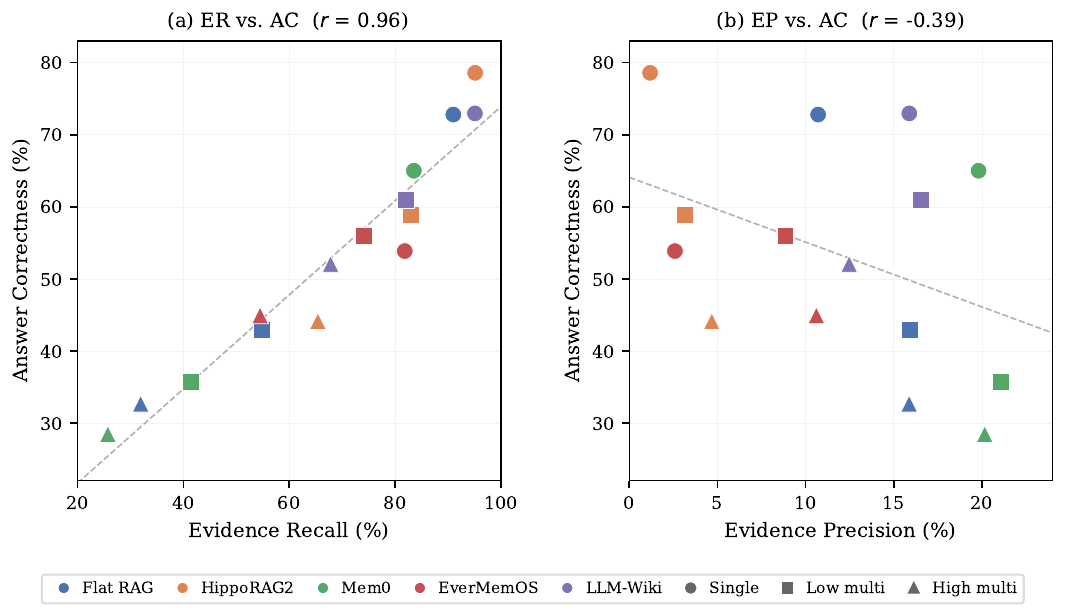}
\caption{Diagnostic scatter plots showing the strong ER--AC correlation ($r{=}0.96$) versus the weak EP--AC relationship ($r{=}{-}0.39$) in the main GLM+Qwen setting. Points represent method--fan-in pairs.}
\label{fig:er-ep-vs-ac}
\end{figure}

\begin{table}[t]
\centering
\small
\begin{tabular}{llrrr}
\toprule
Setting & Method & AC & ER & EP \\
\midrule
\multirow{8}{*}{Single-doc} & Closed-book & 6.5 & -- & -- \\
 & Long-context & 28.0 & -- & -- \\
 & Oracle & 81.6 & 100.0 & 100.0 \\
 & Flat RAG & 72.8 & 91.0 & 10.7 \\
 & HippoRAG2 & \best{78.6} & \best{95.1} & 1.2 \\
 & Mem0 & 65.0 & 83.5 & \best{19.8} \\
 & EverMemOS & 53.9 & 81.8 & 2.6 \\
 & LLM-Wiki & 73.0 & \best{95.1} & 15.9 \\
\midrule
\multirow{8}{*}{Low multi-doc} & Closed-book & 16.4 & -- & -- \\
 & Long-context & 28.0 & -- & -- \\
 & Oracle & 71.1 & 100.0 & 100.0 \\
 & Flat RAG & 43.0 & 54.9 & 15.9 \\
 & HippoRAG2 & 58.9 & \best{83.0} & 3.2 \\
 & Mem0 & 35.8 & 41.4 & \best{21.1} \\
 & EverMemOS & 56.0 & 74.1 & 8.9 \\
 & LLM-Wiki & \best{60.9} & 82.0 & 16.5 \\
\midrule
\multirow{8}{*}{High multi-doc} & Closed-book & 19.0 & -- & -- \\
 & Long-context & 27.4 & -- & -- \\
 & Oracle & 69.8 & 100.0 & 100.0 \\
 & Flat RAG & 32.7 & 31.9 & 15.9 \\
 & HippoRAG2 & 44.1 & 65.4 & 4.7 \\
 & Mem0 & 28.5 & 25.7 & \best{20.2} \\
 & EverMemOS & 45.0 & 54.5 & 10.6 \\
 & LLM-Wiki & \best{52.1} & \best{67.8} & 12.5 \\
\midrule
\multirow{8}{*}{All} & Closed-book & 10.7 & -- & -- \\
 & Long-context & 27.9 & -- & -- \\
 & Oracle & 77.4 & 100.0 & 100.0 \\
 & Flat RAG & 59.9 & 74.1 & 12.7 \\
 & HippoRAG2 & \best{69.1} & 88.3 & 2.1 \\
 & Mem0 & 52.8 & 65.4 & \best{20.2} \\
 & EverMemOS & 53.3 & 76.4 & 5.2 \\
 & LLM-Wiki & 67.5 & \best{88.7} & 15.7 \\
\bottomrule
\end{tabular}
\caption{Main results under the GLM+Qwen setting. AC: answer correctness; ER: evidence recall; EP: evidence precision. Closed-book and long-context baselines do not produce evidence packs for ER/EP evaluation. Oracle feeds the annotated gold evidence to the same QA model, providing a gold-evidence reference condition (ER and EP are 100 by definition). Bold indicates the best non-oracle evidence-construction system within each fan-in block and metric. Table~\ref{tab:collapse-patterns} maps each system to its diagnostic degradation pattern.}
\label{tab:main-glm-qwen}
\end{table}

Table~\ref{tab:main-glm-qwen} shows that the main GLM+Qwen setting is recall-dominant, and Figure~\ref{fig:er-ep-vs-ac} visualizes the corresponding ER--AC and EP--AC relationships. Systems with higher ER generally achieve higher AC even when their EP is low. HippoRAG2 is the clearest example: it retrieves noisy packs with EP of only 2.1\%, yet its high ER (88.3\%) yields the highest non-oracle downstream AC (69.1). Conversely, Mem0 achieves the highest EP (20.2\%) but the lowest non-baseline AC (52.8), as its compressed memory entries omit substantial required support (ER = 65.4\%). LLM-Wiki combines the highest overall ER (88.7\%) with moderate EP (15.7\%), achieving the second-highest non-oracle AC (67.5).

This pattern identifies evidence sufficiency as the main observed bottleneck in the primary setting. In thematically dense corpora, retrieval systems frequently include near-miss passages that share the author's style and topic. Under the evaluated reader and context budgets, this noise is less damaging than omitting required support.

A failure attribution analysis further supports this conclusion. Among instances where systems produce incorrect answers (AC $<$ 33\%), the vast majority co-occur with low ER: 86\% of Flat RAG failures, 88\% of Mem0 failures, and 79\% of HippoRAG2 failures involve ER below 50\%. The Flat RAG top-$k$ ablation provides a controlled intervention: increasing $k$ from 5 to 15 adds retrieval noise (EP drops from 20.8\% to 9.9\%) while simultaneously raising both ER (67.2\% $\to$ 76.7\%) and AC (54.8 $\to$ 61.4). Enlarging the evidence pack therefore improves downstream correctness despite reducing precision, consistent with a recall-dominated regime.

This does not mean precision is unconditionally irrelevant. The recall-dominance result is conditioned on the main setting's context budget and reader model. Section~\ref{sec:model-family-ablation} shows that EP becomes diagnostically meaningful once fan-in increases or the reader changes---precision captures evidence \emph{usability} rather than sufficiency, and its predictive value is setting-dependent.

\subsection{Model-Family Ablation}
\label{sec:model-family-ablation}

\begin{figure}[t]
\centering
\includegraphics[width=\columnwidth]{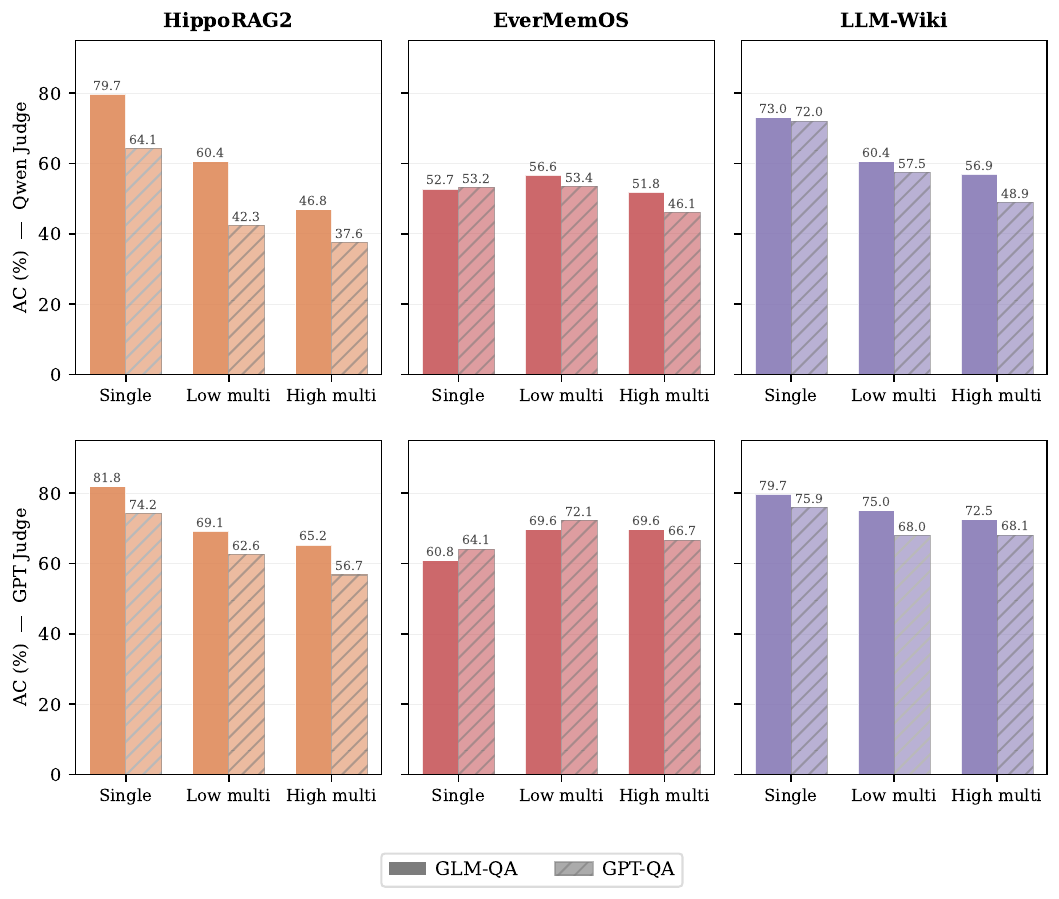}
\caption{Model-family ablation (20\% stratified sample, $n{=}419$). Each panel compares GLM-5.1-FP8 vs.\ GPT-4.1-mini as QA reader on held-fixed evidence packs, controlling for judge identity. Top row: Qwen judge; bottom row: GPT judge.}
\label{fig:model-ablation}
\end{figure}

To test whether the diagnostic patterns depend on a particular model family, we run a stratified ablation on a 20\% sample (419 instances) of AuthTrace. Evidence packs are held fixed; we cross QA reader (GLM-5.1-FP8 vs.\ GPT-4.1-mini) with judge (Qwen3.5-397B-A17B vs.\ GPT-4o-mini), yielding four conditions per system. Figure~\ref{fig:model-ablation} separates QA-model effects from judge-calibration effects by comparing readers under the same judge.

Figure~\ref{fig:model-ablation} shows a clear interaction between QA model and evidence format. With GLM-5.1-FP8 as reader, HippoRAG2's high-recall raw evidence is strongest on single-doc questions. With GPT-4.1-mini, EverMemOS significantly surpasses HippoRAG2 on multi-doc settings (72.1 vs.\ 62.6 at low-multi, $p < 0.01$; 66.7 vs.\ 56.7 at high-multi, $p < 0.05$; both under GPT judge), and LLM-Wiki achieves the highest high-fan-in AC (68.1, significantly above HippoRAG2's 56.7, $p < 0.01$).

The ablation exposes a nuanced role for EP that depends on both fan-in and reader. In the main GLM+Qwen single-doc setting, HippoRAG2 and LLM-Wiki share identical ER (95.1\%), yet HippoRAG2 achieves higher AC (78.6 vs.\ 73.0) despite EP of only 1.2\% versus 15.9\%. One plausible explanation is that HippoRAG2's full-document retrieval provides surrounding local context that aids interpretation of the gold evidence, even though this context does not itself match gold annotations. In the GPT+GPT high-fan-in split, however, LLM-Wiki and HippoRAG2 again share nearly identical ER (70.6\% vs.\ 70.4\%), but now LLM-Wiki's higher EP (13.9\% vs.\ 5.2\%) corresponds to 11.4pp higher AC (68.1 vs.\ 56.7). When evidence spans many documents, the full-document packs that may help locally instead increase synthesis burden, and GPT-4.1-mini benefits from focused, pre-organized evidence.

Together, these observations refine the diagnostic roles of ER and EP. ER captures evidence \emph{sufficiency}---the binding constraint whenever the reader has spare capacity to tolerate noise. EP captures evidence \emph{usability}---its predictive value emerges when synthesis complexity rises or the reader benefits from focused input. For local grounding under generous budgets, surrounding context can compensate for low EP; for broad multi-document synthesis or capacity-limited readers, high-density structured packs reduce synthesis cost. The two metrics are therefore complementary diagnostics whose relative importance shifts with fan-in and reader characteristics, rather than universally ordered priorities.

\subsection{Fan-in Reveals Structural Collapse Patterns}

Fan-in separates systems that appear similar under local grounding by exposing their distinct degradation mechanisms as required source documents increase (Figure~\ref{fig:fanin-degradation}). Flat RAG and Mem0 exhibit cliff-like degradation ($-$55\% and $-$56\% relative degradation): local chunks and independent memory entries have limited cross-document assembly capacity, so performance drops sharply once the query exceeds single-document scope. HippoRAG2 follows a stepwise pattern ($-$44\%): graph traversal extends evidence reach beyond local semantic similarity, but its effective radius is bounded, and performance declines substantially once fan-in exceeds 2--3 documents. LLM-Wiki degrades gradually ($-$29\%): thematic pre-organization provides natural cross-document coverage, trading single-document detail for robust multi-document assembly. EverMemOS degrades the least overall ($-$17\%): its event-level abstraction provides a middle ground between local chunks and full thematic organization, maintaining coverage with less explicit cross-document assembly.

The main-table differences between the best system and runner-up at single-doc and high-multi fan-in are statistically significant by paired bootstrap; the low-multi gap is directional but not significant. Appendix~\ref{app:additional-analyses} gives confidence intervals and per-author robustness checks.

The oracle setting provides a synthesis-difficulty reference: even with annotated gold evidence, AC drops from 81.6 (single-doc) to 69.8 (high-multi), a $-$14.5\% decline attributable to increased synthesis difficulty under the same reader. Subtracting this reference, Flat RAG and Mem0 incur 41--42pp of \emph{additional} construction loss, while LLM-Wiki adds only 14pp, indicating that its organization layer preserves robustness as fan-in increases.

Notably, the different paradigms fail on largely non-overlapping instance sets: the Jaccard overlap of failure sets (AC $<$ 33\%) ranges from 0.04 to 0.23 across system pairs. This complementary error structure suggests that evidence construction quality can be further improved by routing queries to paradigm-appropriate systems based on estimated fan-in.

\subsection{Effective Retrieval Radius and System Selection}

The fan-in gradient enables a fine-grained view of when each system's advantage begins and ends. Table~\ref{tab:crossover} reports AC at each exact fan-in level for the three most competitive systems. HippoRAG2 leads significantly at fan-in$\,{=}\,$1 ($+$5.6pp, $p{<}0.01$); the two systems are statistically indistinguishable at fan-in$\,{=}\,$2, and LLM-Wiki pulls ahead significantly at fan-in 3--4 ($+$8--9pp, $p{<}0.05$). This crossover characterizes an \emph{effective retrieval radius} for each paradigm: graph traversal is most effective within a single document's semantic neighborhood, while thematic pre-organization becomes advantageous once multi-document aggregation is needed.

\begin{table}[t]
\centering
\small
\setlength{\tabcolsep}{4pt}
\begin{tabular}{lccccc}
\toprule
Method & $f{=}1$ & $f{=}2$ & $f{=}3$ & $f{=}4$ & $f{=}5$ \\
$n$ & 1285 & 552 & 72 & 102 & 79 \\
\midrule
HippoRAG2 & \best{78.6}$^\dagger$ & 58.9 & 47.2 & 43.8 & 43.5 \\
LLM-Wiki & 73.0 & \best{60.9} & \best{55.6}$^\dagger$ & \best{52.4}$^\dagger$ & \best{48.1} \\
EverMemOS & 53.9 & 56.0 & 47.7 & 47.7 & 40.6 \\
\bottomrule
\end{tabular}
\caption{AC at each exact fan-in level ($f$). $n$: number of benchmark instances per level (9 additional instances at $f{\geq}6$ are omitted due to small sample size). Bold marks the best system; $^\dagger$ indicates the lead over the runner-up is statistically significant (paired bootstrap, $p{<}0.05$, 10K resamples).}
\label{tab:crossover}
\end{table}

EverMemOS provides a contrasting profile: it maintains relatively stable AC across fan-in levels (53.9 at $f{=}1$, 56.0 at $f{=}2$, declining to 40.6 at $f{=}5$), with lower peak performance but the slowest overall degradation among systems that do not employ thematic pre-organization.

This pattern suggests a simple routing rule: graph-based retrieval is preferable for local detail queries, while structured thematic indexing is more stable for mixed or predominantly multi-document workloads ($f \geq 2$). Appendix~\ref{app:case-studies} provides two contrastive failure cases that make the underlying mechanisms directly observable: a rare-entity question that graph retrieval answers correctly but thematic abstraction hallucinates, and a five-document synthesis that thematic search covers completely but graph traversal cannot reach.

By contrast, budgeted long-context prompting (150K-character sequential corpus slice) achieves only 27.4--28.0 AC across all fan-in levels---below even Flat RAG at high fan-in---confirming that unstructured raw exposure is not a substitute for query-targeted evidence construction \citep{bai-etal-2024-longbench,hsieh2024ruler}.

\section{Related Work}

\paragraph{Cross-paradigm evaluation of evidence construction.}
Prior evaluations of evidence-construction paradigms remain siloed by design. RAG evaluation focuses on answer faithfulness, context relevance, and citation quality \citep{es-etal-2024-ragas,saad-falcon-etal-2024-ares,gao-etal-2023-enabling}; multi-document QA benchmarks test reasoning over distributed sources \citep{yang-etal-2018-hotpotqa,ho-etal-2020-constructing,trivedi-etal-2022-musique}; long-context benchmarks evaluate whether models can use information already placed in extended inputs \citep{kocisky-etal-2018-narrativeqa,bai-etal-2024-longbench,hsieh2024ruler,zhang-etal-2024-bench}; memory benchmarks study storage and recall over long interaction histories \citep{maharana-etal-2024-evaluating,ICLR2025_d813d324}; and graph or hierarchical retrieval methods are evaluated through their own retrieval and summarization protocols \citep{edge2025localglobalgraphrag,NEURIPS2024_6ddc001d,pmlr-v267-gutierrez25a,sarthi2024raptor}. Because these evaluations differ in corpus assumptions and evidence interfaces, they make it difficult to diagnose under a single protocol which evidence-organization paradigm fails, where it fails, or whether the failure comes from missing evidence, noisy evidence, or answer synthesis. AuthTrace is designed as a shared testbed that places chunk retrieval, memory systems, graph retrieval, thematic indexing, and long-context prompting on the same corpus, query set, and evidence-pack protocol.

\paragraph{Existing paradigm-specific benchmarks.}
Paradigm-specific benchmarks test reasoning within a single evidence interface. Multi-hop and multi-document QA datasets such as HotpotQA, 2WikiMultiHopQA, and MuSiQue emphasize bridge entities, compositional reasoning, and explicit supporting facts across heterogeneous sources \citep{yang-etal-2018-hotpotqa,ho-etal-2020-constructing,trivedi-etal-2022-musique}. Long-form and many-answer QA datasets such as ASQA, ELI5, and QAMPARI further stress retrieval and synthesis over multiple pieces of evidence \citep{stelmakh-etal-2022-asqa,fan-etal-2019-eli5,amouyal-etal-2023-qampari}. These benchmarks are valuable for testing answer synthesis and support attribution, but they do not isolate evidence-pack construction across competing organization paradigms. Long-context benchmarks such as NarrativeQA, LongBench, RULER, and $\infty$Bench test reader-side use of long inputs \citep{kocisky-etal-2018-narrativeqa,bai-etal-2024-longbench,hsieh2024ruler,zhang-etal-2024-bench}, but they typically assume the relevant information is already present in the prompt. Agent-memory systems and benchmarks, including MemGPT, Mem0, A-MEM, LoCoMo, and LongMemEval, study how information is stored, compressed, updated, and retrieved over time \citep{packer2024memgptllmsoperatingsystems,DBLP:conf/ecai/ChhikaraKASY25,xu2025amemagenticmemoryllm,maharana-etal-2024-evaluating,ICLR2025_d813d324}. AuthTrace complements these lines of work by treating a single-author corpus as a controlled long-term knowledge source and asking whether different organization layers preserve the concrete evidence needed for grounded answering under increasing fan-in.

\paragraph{Evidence versus answer evaluation.}
Answer-level metrics conflate evidence quality with generation quality. RAGAS evaluates RAG pipelines through dimensions such as faithfulness and context relevance \citep{es-etal-2024-ragas}; ARES automates RAG evaluation with model-based judges and prediction-powered inference \citep{saad-falcon-etal-2024-ares}; and ALCE evaluates citation quality for long-form answers \citep{gao-etal-2023-enabling}. GraphRAG, HippoRAG, HippoRAG2, and RAPTOR show that graph, memory, and hierarchical structures can improve retrieval beyond flat chunk matching \citep{edge2025localglobalgraphrag,NEURIPS2024_6ddc001d,pmlr-v267-gutierrez25a,sarthi2024raptor}, but their evaluations do not directly separate evidence sufficiency from downstream answer generation. AuthTrace positions evidence construction as a distinct object of evaluation. It measures whether the predicted evidence pack covers all required support through evidence recall, whether it avoids topically plausible but non-supporting passages through evidence precision, and how these evidence-stage metrics explain answer correctness. This makes AuthTrace complementary to answer- and citation-level evaluation while enabling cross-paradigm diagnosis on a unified protocol.

\section{Conclusion}

We present AuthTrace, a diagnostic benchmark that treats single-author corpora as controlled evidence sources, enabling unified evaluation of retrieval, memory, graph, and structured-evidence systems under a shared evidence protocol.

Evaluating eight systems reveals three insights. No paradigm dominates: graph retrieval excels at single-document grounding while thematic organization leads on multi-document synthesis, with a crossover at fan-in 2. Evidence recall is the strongest observed evidence-stage predictor of answer correctness in the main setting ($r = 0.96$); most failures stem from missing evidence rather than answer synthesis alone. Fixed-order long-context prompting performs weakly across fan-in levels, showing that budgeted raw corpus exposure is not a substitute for query-targeted evidence construction.

Rather than producing a single ranking, AuthTrace identifies where each paradigm fails, why, and which workload it best serves (Appendix~\ref{app:implications} distills actionable selection guidelines). More broadly, fan-in decomposition offers a transferable diagnostic axis: any domain with thematically dense, overlapping sources—creator archives, institutional repositories, or long-running research programs—can adopt the same protocol to expose paradigm-specific failure modes. 

\section*{Limitations}

AuthTrace focuses on modern Chinese prose corpora from five modern authors. This setting provides a controlled and reproducible proxy for creator-centered knowledge bases, and broader domains can further test how the diagnostic patterns transfer to newsletters, blogs, institutional archives, and technical writing. The benchmark uses LLM-assisted instance construction and model-based evaluation, mitigated by automated filtering and human spot-checking \citep{saad-falcon-etal-2024-ares,es-etal-2024-ragas,liu-etal-2023-g,NEURIPS2023_91f18a12}. Future extensions can broaden the author set, vary context budgets to study when evidence precision becomes predictive, and evaluate adaptive systems that route among multiple evidence representations. Because paradigms naturally expose evidence at different granularities, AuthTrace evaluates them through a normalized external contract rather than identical internal operating conditions.

Our long-context baseline is limited by the reader's finite input window. We therefore use a fixed-order 150K-character slice of each author corpus rather than the full corpus. This means that, for some questions, the gold source documents may fall outside the provided context. The baseline should therefore be interpreted as measuring budgeted fixed-order raw-context exposure, not as an optimized or full-corpus long-context upper bound. Future work can revisit this comparison with models that support full-corpus inputs or with controlled variants that guarantee gold-document inclusion while preserving large distractor contexts.

\section*{Ethics Statement}

AuthTrace is built from published, publicly available writings by modern Chinese-language authors. The selected authors all died more than fifty years before dataset construction; under the general copyright term for natural-person works in the Copyright Law of the People's Republic of China, economic rights in natural-person works generally last for the author's lifetime and fifty years after death, ending on December 31 of the fiftieth year. On this basis, we treat the source materials as public-domain materials in the relevant source context. We nevertheless avoid making a universal copyright claim across jurisdictions, and release metadata should preserve source provenance and author attribution.

The benchmark is intended for evaluating evidence selection, evidence assembly, and grounded answering, rather than for replacing scholarly reading of the original texts. Released materials should document source provenance, preprocessing decisions, filtering criteria, annotation procedures, and evaluation rubrics. Redistribution of any underlying source texts should follow the copyright rules applicable in the relevant jurisdiction; released provenance metadata is intended to help users verify the status of sources in their own use cases.

\bibliography{custom}

\appendix

\section{Additional Task Motivation}
\label{app:task-details}

\subsection{Why Thematically Dense Single-Author Corpora Are Diagnostic}

Single-author corpora create dense semantic neighborhoods. Documents share voice, vocabulary, recurring topics, historical context, and rhetorical habits. As a result, the nearest passages to a query are often plausible distractors rather than direct support. This makes the corpus a natural stress test for evidence construction: a useful system must distinguish evidence from thematic adjacency.

The setting also captures a common real-world use case. Users often query a creator-centered knowledge base: a columnist's archive, a scholar's essays, a blogger's posts, a novelist's public writings, or a long-running newsletter. Such archives contain repeated arguments, evolving positions, and recurring examples. Many questions therefore require a compact evidence pack that combines local textual detail with cross-document synthesis.

\subsection{A Unified Testbed for Retrieval and Memory Paradigms}

A single-author corpus occupies a useful middle ground between open-domain document retrieval and personal memory. From a retrieval perspective, it is a bounded document collection that can be chunked, embedded, and searched. From a memory perspective, it resembles an accumulated knowledge source and can be ingested, compressed, organized, and retrieved as long-term memory. From a graph perspective, recurring people, places, concepts, events, and claims form relational structure. From a wiki perspective, repeated themes can be organized into structured pages with provenance.

This property makes AuthTrace a useful diagnostic testbed. All systems start from the same raw corpus and answer the same questions, but each system imposes a different organization layer: flat chunks, memory entries, graph neighborhoods, event abstractions, thematic pages, or raw long context. The resulting comparison emphasizes representation choices rather than dataset mismatch. Table~\ref{tab:paradigms} summarizes the corpus view and evidence-construction mechanism associated with each paradigm.

\begin{table}[t]
\centering
\small
\begin{tabularx}{\columnwidth}{@{}lXX@{}}
\toprule
Paradigm & Corpus view & Evidence-construction view \\
\midrule
Flat RAG & Document collection & Chunk retrieval and ranking \\
Memory & Accumulated knowledge & Ingest, compression, recall \\
Graph RAG & Relational structure & Entity/relation traversal \\
Event memory & Temporal experience & Event abstraction and retrieval \\
LLM-Wiki & Thematic archive & Wiki compilation and traversal \\
Long-context & Sequential corpus slice & Reader-side selection \\
\bottomrule
\end{tabularx}
\caption{AuthTrace supports cross-paradigm diagnosis because a single-author corpus can be instantiated as retrieval, memory, graph, event, wiki, and long-context views under the same query set and annotations.}
\label{tab:paradigms}
\end{table}

\section{Dataset Generation and Annotation Pipeline}
\label{app:dataset-pipeline}

This appendix describes the full construction pipeline used to create AuthTrace. The main design goal is to make evidence requirements explicit before evaluation: each accepted instance contains a bounded query, quoted source evidence required by the reference answer, atomic claim units, a reference answer, and an exact document-level fan-in label.

\subsection{Corpus Filtering and Document Cleaning}

We begin from public-domain writings by five modern Chinese prose writers. The raw collection contains essays, criticism, memoir, travel writing, fiction, and other prose genres. Since AuthTrace evaluates author-grounded evidence construction, the document filter emphasizes texts where the author's own stance, observation, or argument is the central object of interpretation. We use Qwen3.5-397B-A17B as an LLM-based document classifier to retain articles that satisfy two criteria: the text is nonfictional prose, and the majority of the article expresses the author's own viewpoint rather than quoted material, editorial framing, or fictional narration. Documents shorter than 800 Chinese characters are removed. The remaining documents are cleaned by normalizing text encoding, whitespace, metadata fields, document identifiers, and title records. This produces the 860-document corpus used in the benchmark. Table~\ref{tab:corpus} reports per-author statistics.

\begin{table}[t]
\centering
\small
\begin{tabular}{lrrr}
\toprule
Author & Articles & Avg. Length & Total Chars \\
\midrule
Lu Xun & 276 & 2,203 & 608,154 \\
Zhou Zuoren & 305 & 1,816 & 553,960 \\
Zhu Ziqing & 136 & 2,985 & 406,004 \\
Xiao Hong & 64 & 2,183 & 139,712 \\
Yu Dafu & 79 & 3,563 & 281,538 \\
\midrule
Total & 860 & 2,313 & 1,989,368 \\
\bottomrule
\end{tabular}
\caption{AuthTrace corpus statistics by author. Avg. Length and Total Chars are measured in Chinese characters after document filtering.}
\label{tab:corpus}
\end{table}

\subsection{Canonical Theme Taxonomy}

To support controlled multi-document sampling, every retained article is labeled with a canonical theme. We use a fixed taxonomy designed to cover literary, social, political, biographical, embodied, emotional, spatial, and educational concerns common in the target corpora. The taxonomy is shown in Table~\ref{tab:canonical-themes}. The same Qwen3.5-397B-A17B labeling pass assigns theme scores, a primary theme, optional secondary themes, a theme-focus score, a QA-suitability score, and a keep flag. The primary theme is used for stratified single-document sampling and same-theme multi-document grouping.

\begin{table}[t]
\centering
\footnotesize
\begin{tabularx}{\columnwidth}{@{}lX@{}}
\toprule
Theme & Scope \\
\midrule
T1: Literary Theory & Literature, composition, genre, criticism \\
T2: Language \& Style & Rhetoric, translation, voice \\
T3: Cultural Critique & Customs, institutions, norms \\
T4: Politics \& War & Conflict, revolution, nationalism \\
T5: Memory \& Biography & Remembrance, portraiture \\
T6: Intellectual Life & Writers, publishing, authorship \\
T7: Gender \& Relations & Family, marriage, social roles \\
T8: Hardship & Poverty, illness, labor, survival \\
T9: Emotion \& Self & Affect, solitude, self-analysis \\
T10: Travel \& Place & Landscape, locality, space \\
T11: Education \& Youth & Schooling, teachers, learning \\
\bottomrule
\end{tabularx}
\caption{Canonical theme taxonomy used for article-level labeling and same-theme multi-document sampling.}
\label{tab:canonical-themes}
\end{table}

\subsection{Single-Document Instance Generation}

Single-document generation targets local grounding. Candidate articles are filtered by the keep flag, a minimum theme-focus score of 5, and a minimum QA-suitability score of 5. Sampling is stratified by primary theme: articles are grouped by theme, sorted by QA suitability, theme focus, length, and filename, and then selected with a deterministic random seed. This creates a balanced set of article-level generation calls rather than concentrating the benchmark in the largest themes.

For each sampled article, Claude Opus 4.6 receives the article text, author identifier, primary theme, document identifier, and title. The title is provided only as internal metadata for the annotator model. The output is constrained to a JSON object containing an instance list. Each single-document instance must be answerable from the current article, must use the author-neutral expression ``the author,'' and must be anchored by concrete article-specific details rather than by a broad topic. The gold evidence units are direct quotations from the article and are normalized to the current document identifier. The default generation setting produces up to three instances per article with temperature 0.2 and a maximum generation budget of 24,000 tokens.

\subsection{Multi-Document Instance Generation}

Multi-document generation targets cross-document evidence construction. For each author and theme, candidate articles are selected using the same keep, theme-focus, and QA-suitability thresholds as the single-document pipeline. The sampler creates two types of same-theme groups. The low-fan-in pass samples groups of five input documents and asks the generator to produce instances whose actual evidence should come from two to three documents. The high-fan-in pass samples groups of eight input documents and asks the generator to produce instances whose actual evidence should come from five or more documents.

These requested fan-in ranges are used only as generation targets, not as released evaluation labels. In practice, LLM-generated instances may remain genuinely cross-document while using fewer distinct quoted source documents than requested, especially after normalization, duplicate-evidence removal, and evidence-support filtering. We therefore recompute the released fan-in from the number of distinct source documents appearing in the accepted gold evidence units. Consequently, the high multi-doc evaluation bin is defined by the post-filtered evidence label ($f \geq 3$), while exact-fan-in results are reported separately.

Group sampling is deterministic under a fixed seed and prevents duplicate document combinations. Within each theme, documents are weighted by QA suitability, theme focus, and prior usage, encouraging high-quality coverage while spreading generation calls across the candidate pool. The default configuration creates 24 low-fan-in groups and 12 high-fan-in groups per theme when enough candidates are available, with up to four instances per low-fan-in group and five instances per high-fan-in group.

Claude Opus 4.6 is used for the multi-document generation pass under the same temperature and maximum-token settings. The multi-document prompt restricts task types to thematic synthesis, contrastive reasoning, and diachronic evolution. It asks for bounded questions whose answers can be verified from quoted evidence, with query details serving to delimit the evidence target rather than to reveal the answer.

\subsection{Normalization, Rejection, and Final Filtering}

Generated outputs are parsed from JSON and normalized before entering the benchmark pool. Evidence units are accepted only when their document identifiers match the allowed input documents for the generation call. Duplicate evidence spans are removed. Gold claims are normalized into non-empty atomic strings. For single-document instances, all evidence is assigned to the source article and the fan-in is set to 1. For multi-document instances, the source documents are inferred directly from the normalized evidence units; instances with evidence from fewer than two documents are rejected from the multi-document pool.

The normalized pool then goes through benchmark-level filtering for query validity, evidence-answer support, and quote--document consistency. Query validity checks remove title leakage, explicit source-range phrasing, and overly broad questions whose answer boundary is unstable. Evidence-answer support verifies that the reference answer is licensed by the quoted evidence and that the gold claim units are covered. Quote--document consistency verifies that every quoted evidence span is aligned with its declared source document. This process yields 2,099 accepted instances from 2,398 generated candidates.

\section{Annotation Schema}
\label{app:annotation}

Each accepted instance is stored with the following fields:

\begin{promptbox}
\small
\texttt{\{}\\
\texttt{~~"author\_id": "...",}\\
\texttt{~~"query": "...",}\\
\texttt{~~"task\_type": "thematic\_synthesis |}\\
\texttt{~~~~contrastive\_reasoning |}\\
\texttt{~~~~diachronic\_evolution |}\\
\texttt{~~~~local\_grounding",}\\
\texttt{~~"fan\_in": 2,}\\
\texttt{~~"generation\_mode": "single | small | large",}\\
\texttt{~~"theme": "...",}\\
\texttt{~~"gold\_source\_docs": ["doc\_id\_1", "doc\_id\_2"],}\\
\texttt{~~"gold\_evidence\_units": [}\\
\texttt{~~~~\{"doc\_id": "doc\_id\_1",}\\
\texttt{~~~~~"text": "quoted source span"\}}\\
\texttt{~~],}\\
\texttt{~~"gold\_claim\_units": ["atomic claim"],}\\
\texttt{~~"reference\_answer": "..."}\\
\texttt{\}}
\end{promptbox}

\noindent The released fan-in is the exact number of distinct documents appearing in \texttt{gold\_evidence\_units}.

\section{Dataset Examples}
\label{app:examples}

We provide two representative instances from AuthTrace to illustrate the annotation structure. All text is in the original Chinese with English translations for reviewer readability. The first is a single-document instance (fan-in = 1) from Yu Dafu; the second is a multi-document instance (fan-in = 3) from Zhou Zuoren. Each figure shows the query, gold evidence units with source document identifiers, gold claim units, and the reference answer.

\begin{figure*}[t]
\centering
\includegraphics[width=\textwidth]{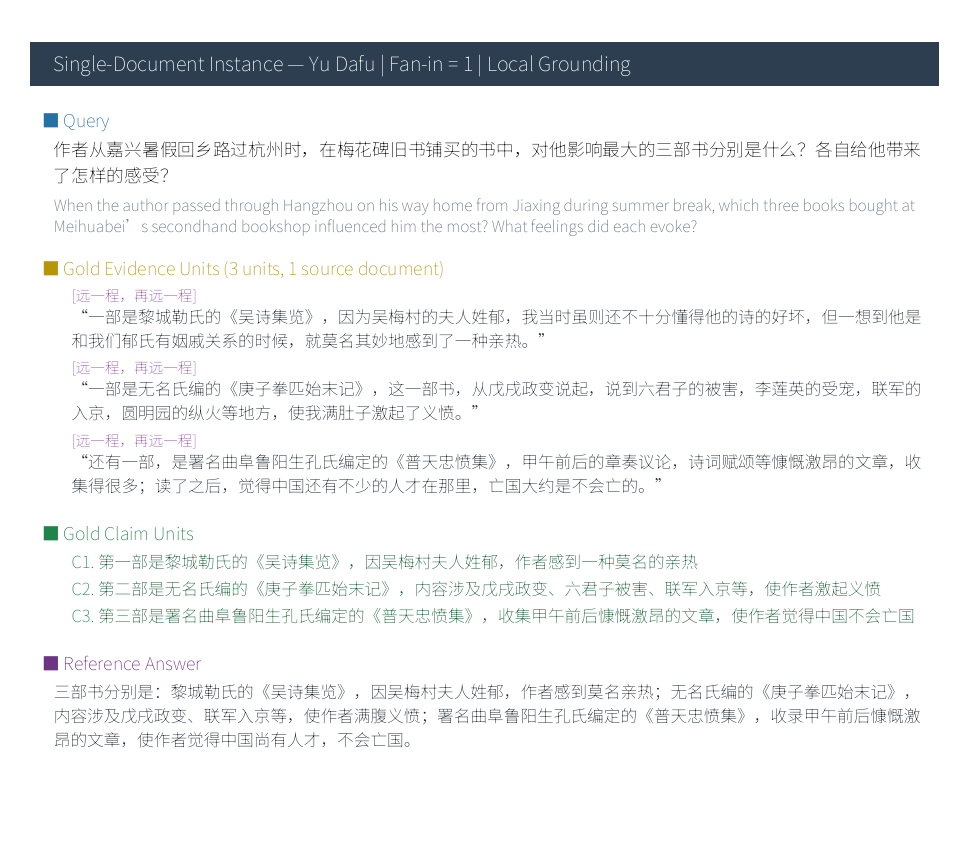}
\caption{Single-document example instance (Yu Dafu, fan-in = 1). This instance tests local grounding: whether a system can locate three specific book descriptions within one memoir essay. The parallel structure (three books, three emotional responses) requires complete extraction. Near-miss passages about Yu Dafu's other reading experiences would be topically plausible but non-supporting.}
\label{fig:example-single}
\end{figure*}

\begin{figure*}[t]
\centering
\includegraphics[width=\textwidth]{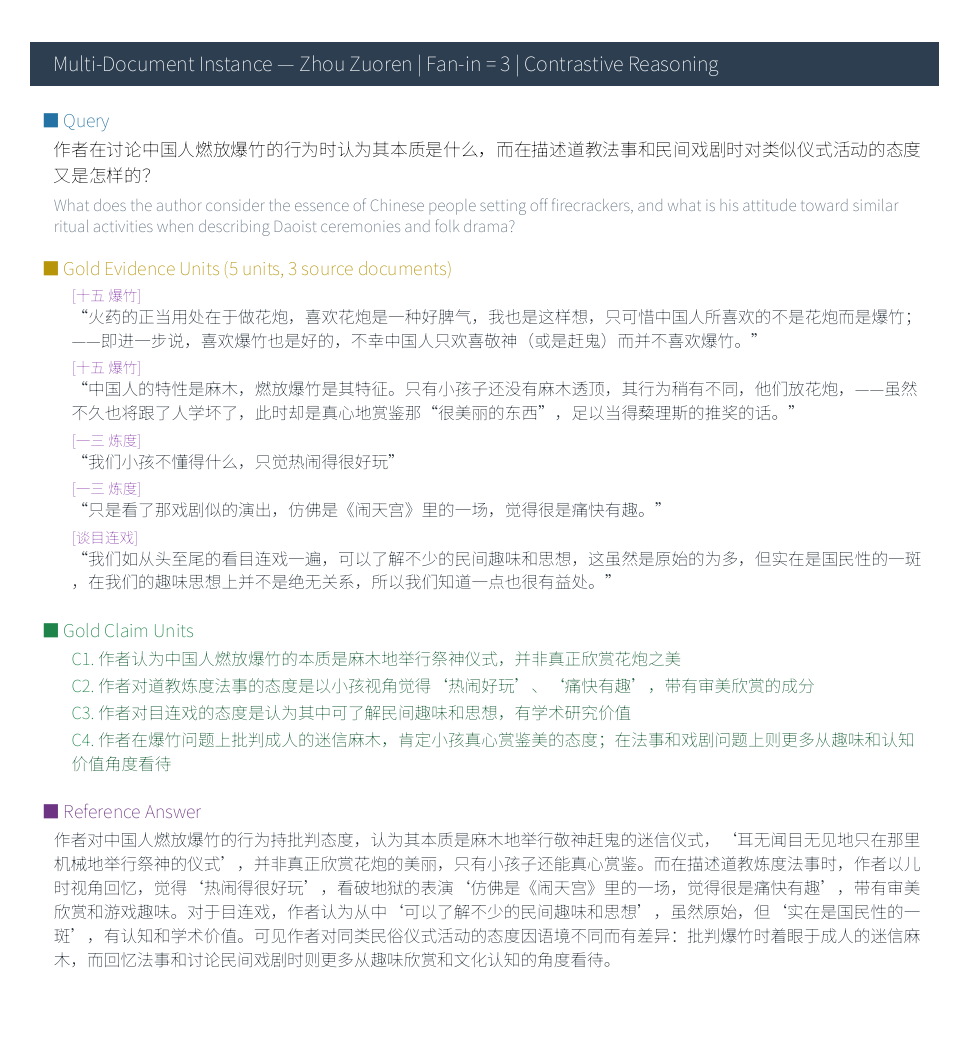}
\caption{Multi-document example instance (Zhou Zuoren, fan-in = 3). This instance tests cross-document synthesis: the system must retrieve evidence from three distinct essays (a cultural critique, a childhood memoir, and a literary commentary) to construct a contrastive answer about the author's differing attitudes toward similar folk rituals. Near-miss passages about Zhou Zuoren's other cultural critiques would be topically plausible but non-supporting.}
\label{fig:example-multi}
\end{figure*}

\section{Metric Interpretation}
\label{app:metric-interpretation}

EP is especially meaningful as a usability and efficiency metric. When the reader has enough context capacity, additional near-miss passages may impose limited cost if the required evidence is present. Under tighter budgets, or when comparing structured representations with different signal densities, the same precision difference can become consequential because irrelevant passages compete with required support and increase synthesis burden. AuthTrace therefore treats EP as diagnostic rather than prescriptive: its downstream effect depends on the reader, the representation, and the context budget.

High ER with low EP indicates over-retrieval with sufficient support. High EP with low ER indicates a clean but incomplete pack. High ER and EP with low AC indicates a synthesis failure rather than an evidence-construction failure. This decomposition lets AuthTrace identify which stage of the pipeline produces the observed answer degradation.

\section{System Descriptions and Implementation Details}
\label{app:implementation}

All systems share the same model infrastructure: GLM-5.1-FP8 \citep{glm5team2026glm5vibecodingagentic} serves as both the index-construction LLM and the QA reader in the main setting (GPT-4.1-mini in the model-family ablation), and Qwen3-Embedding-8B (4096 dimensions) \citep{yang2025qwen3technicalreport} provides embeddings for all retrieval-based systems. EverMemOS additionally uses Qwen3-Reranker-4B for reranking. All systems use the same answer-generation prompt (Section~\ref{app:qa-prompt}). Table~\ref{tab:system-params} summarizes key parameters; detailed descriptions follow.

\begin{table}[t]
\centering
\footnotesize
\setlength{\tabcolsep}{2pt}
\begin{tabular}{lllll}
\toprule
System & Granularity & Top-$k$ & Emb. & LLM \\
\midrule
Closed-book & --- & --- & --- & --- \\
Long-context & Seq. docs & --- & --- & --- \\
Oracle & Gold units & --- & --- & --- \\
Flat RAG & 300-char & 10 & Q3-8B & --- \\
HippoRAG2 & Full docs & 10 & Q3-8B & GLM \\
Mem0 & Mem.\ entries & 10 & Q3-8B & GLM \\
EverMemOS & Episodes & 10 & Q3-8B & GLM \\
LLM-Wiki & Wiki pages & 10 & Q3-8B & GLM \\
\bottomrule
\end{tabular}
\caption{System parameters. Q3-8B = Qwen3-Embedding-8B (4096-d). GLM = GLM-5.1-FP8. All systems use GLM-5.1-FP8 as QA reader (temp.\ 0.0). EverMemOS uses Qwen3-Reranker-4B for reranking.}
\label{tab:system-params}
\end{table}

\paragraph{Closed-book.}
The model receives only the query and a background hint specifying the author's name. No context is provided. This baseline estimates how much can be answered from parametric knowledge or broad literary priors.

\paragraph{Long-context.}
The model receives the query together with a sequentially concatenated slice of the corresponding author corpus. Documents are sorted alphabetically by filename and concatenated with their titles prepended; documents are included in order until a 150K-character input budget is reached. We choose this budget as a conservative operating point under the deployed GLM-5.1/GLM-5.1-FP8 200K-token context window, accounting for tokenizer variation on Chinese text and reserving space for instructions and generation. This tests whether large-budget raw corpus exposure can provide the evidence needed for answering without an explicit evidence-construction stage.

\paragraph{Oracle evidence.}
The model receives the gold evidence units concatenated as context. This provides a gold-evidence reference condition for answer generation from the annotated evidence. The gold evidence is formatted identically to predicted evidence packs.

\paragraph{Flat RAG.}
The corpus is segmented into chunks of 300 Chinese characters with 10-character overlap, embedded with Qwen3-Embedding-8B, and retrieved via cosine similarity without reranking. At query time, the top-$k$ chunks are returned (default $k{=}10$; ablated at $k{=}5$ and $k{=}15$) and passed directly to the answer model.

\paragraph{HippoRAG2.}
HippoRAG2 constructs a knowledge graph via OpenIE and performs multi-step graph traversal at retrieval time. We use GLM-5.1-FP8 for graph construction, Qwen3-Embedding-8B for embeddings, up to 3 traversal steps with a retrieval pool of 200 candidates and linking top-$k$ of 5. The top 10 retrieved documents (full articles) are passed to the QA reader. Answer generation uses a maximum of 2,048 tokens at temperature 0.0.

\paragraph{Mem0.}
Mem0 extracts and compresses memory entries from the chunked corpus (300-character chunks, matching Flat RAG) using GLM-5.1-FP8, producing abstracted entries shorter than raw chunks. We use Qwen3-Embedding-8B for embeddings. At query time, hybrid search retrieves the top 10 memory entries for the QA reader.

\paragraph{EverMemOS.}
EverMemOS performs event-oriented memory extraction over episodic memory, with agentic retrieval and reranking. We use GLM-5.1-FP8 for memory extraction, Qwen3-Embedding-8B for embeddings, and Qwen3-Reranker-4B for reranking. At query time, the top 10 episodic memory units are passed to the QA reader. Answer generation uses a maximum of 2,048 tokens at temperature 0.0.

\paragraph{LLM-Wiki evidence construction.}
LLM-Wiki \citep{ming2026retrievalreasoningselfevolvingagentnative} compiles raw documents into structured, interlinked Wiki pages organized by thematic directories, with bidirectional wikilinks and source provenance, and uses an Error Book mechanism for persistent self-correction of compilation errors. At query time, an agent performs compositional retrieval by iteratively calling \texttt{wiki\_search} and \texttt{wiki\_read} to navigate the Wiki structure until sufficient evidence is gathered. We use GLM-5.1-FP8 for Wiki compilation, Qwen3-Embedding-8B for the search index, a maximum tool-call budget of $T_{\max}{=}15$, and patience threshold $P{=}3$. The top 10 evidence segments from the agent's traversal are passed to the QA reader. Answer generation uses a maximum of 2,048 tokens at temperature 0.0.

\paragraph{Shared cross-paradigm comparison.}
AuthTrace compares these systems through a shared external contract. Each system ingests the same cleaned author corpus, receives the same query, outputs an evidence pack when applicable, and is judged against the same evidence and claim annotations. All systems use the same QA reader model and the same answer-generation prompt, so measured differences in AC primarily reflect the evidence made available to the reader rather than differences in the reader itself.

\subsection{Answer Generation Prompt}
\label{app:qa-prompt}

All systems that produce evidence packs use the same QA prompt for answer generation. The original prompt is in Chinese; we provide the English translation below.

\promptheading{System prompt.}

\begin{promptbox}
\small
\texttt{You are a Chinese prose question-answering assistant. Please answer the question based on the given context; if the context is insufficient to support the answer, explicitly state that the evidence is insufficient.}
\end{promptbox}

\promptheading{User prompt template.}

\begin{promptbox}
\small
\texttt{Please answer the question.}

\texttt{Requirements:}\\
\texttt{1. Answer based on the context; do not fabricate.}\\
\texttt{2. The answer should completely cover the key points required by the question.}\\
\texttt{3. If the context is insufficient to support the answer, state that the evidence is insufficient.}\\
\texttt{4. Do not output your reasoning process.}

\texttt{[Question] \{query\}}

\texttt{[Context]}\\
\texttt{[Evidence 1] \{text\_1\}}\\
\texttt{[Evidence 2] \{text\_2\}}\\
\texttt{...}

\texttt{Please provide the answer directly.}
\end{promptbox}

For the closed-book setting, the context section is replaced with: ``[Please answer based on prior knowledge and the prompt directly]'', and an additional line specifies the author's name.

\section{Additional Diagnostic Analyses}
\label{app:additional-analyses}

\subsection{Flat RAG Retrieval-Budget Ablation}

We further vary the number of retrieved chunks for Flat RAG. Increasing top-$k$ expands the evidence pack and therefore creates a direct test of the recall--precision tradeoff. As top-$k$ increases from 5 to 10 to 15, EP decreases because the retrieved pack contains more thematically plausible but non-supporting passages. At the same time, AC continues to improve, indicating that the reader benefits from the additional coverage before context noise becomes the dominant bottleneck.

\subsection{Statistical Significance and Per-Author Robustness}

The main-table differences between the best system and runner-up are statistically significant by paired bootstrap (10,000 resamples, percentile 95\% CI; $p < 0.05$) at single-doc and high-multi fan-in: HippoRAG2 vs.\ LLM-Wiki at single-doc (+5.6pp, $p < 0.01$) and LLM-Wiki vs.\ EverMemOS at high-multi (+7.1pp, $p < 0.01$). The low-multi difference between LLM-Wiki and HippoRAG2 (+2.0pp) is not statistically significant (95\% CI includes zero). The ER--AC correlation ($r = 0.96$) has a bootstrap 95\% CI of [0.92, 0.99]; the EP--AC correlation ($-$0.39) has a CI that includes zero, confirming EP's limited predictive role in the main setting.

The main trends are also stable across authors. The ER--AC correlation exceeds 0.92 for every individual author corpus (range: 0.930--0.962), and the per-author best system consistently follows the fan-in-dependent pattern: HippoRAG2 leads single-doc for all five authors, while LLM-Wiki leads high-multi for all five. The diagnostic conclusions are therefore not driven by any single author corpus.

\subsection{Oracle-Normalized Evidence-Construction Loss}

The oracle setting also degrades from single-doc to high multi-doc questions, even when gold evidence is provided. This degradation reflects the increased synthesis difficulty of high-fan-in questions under the same reader. We therefore use oracle degradation as a diagnostic reference and subtract it from each system's degradation to estimate the additional loss associated with evidence construction.

\begin{table}[t]
\centering
\small
\setlength{\tabcolsep}{3pt}
\begin{tabular}{lrrrr}
\toprule
Method & Single AC & High AC & Rel.\ Drop & Oracle-norm. \\
\midrule
Oracle & 81.6 & 69.8 & $-$14.5\% & -- \\
HippoRAG2 & 78.6 & 44.1 & $-$43.9\% & $-$29.4pp \\
Flat RAG & 72.8 & 32.7 & $-$55.1\% & $-$40.6pp \\
Mem0 & 65.0 & 28.5 & $-$56.2\% & $-$41.7pp \\
EverMemOS & 53.9 & 45.0 & $-$16.5\% & $-$2.0pp \\
LLM-Wiki & 73.0 & 52.1 & $-$28.6\% & $-$14.1pp \\
Long-context & 28.0 & 27.4 & $-$2.1\% & -- \\
\bottomrule
\end{tabular}
\caption{Fan-in degradation analysis. Rel.\ Drop = relative AC change from single-doc to high multi-doc. Oracle-norm.\ = additional degradation beyond the oracle reference ($-$14.5\%), separating evidence-construction loss from the observed synthesis-difficulty trend.}
\label{tab:degradation}
\end{table}

This view separates the observed synthesis-difficulty trend from construction failure. LLM-Wiki evidence construction and EverMemOS degrade only slightly beyond the oracle reference, indicating that their organization layers preserve robustness as fan-in increases. Flat RAG and Mem0 incur much larger additional losses, indicating that their main bottleneck is assembling the required evidence rather than generating from it.

\subsection{Collapse Pattern Summary}

\begin{table}[t]
\centering
\small
\setlength{\tabcolsep}{3pt}
\begin{tabularx}{\columnwidth}{@{}p{1.2cm}p{1.8cm}X@{}}
\toprule
Pattern & Methods & Interpretation \\
\midrule
Cliff-like & Flat RAG, Mem0 & Local retrieval and memory compression lack a broad assembly mechanism. \\
Step-like & HippoRAG2 & Graph traversal extends retrieval until signal decays beyond 2--3 hops. \\
Gradual & LLM-Wiki & Thematic pre-aggregation supports broad synthesis at the cost of local detail. \\
Rebound & EverMemOS & Event granularity aligns with moderate cross-document aggregation. \\
\bottomrule
\end{tabularx}
\caption{Fan-in reveals structural degradation profiles. Each pattern corresponds to a different evidence-organization granularity.}
\label{tab:collapse-patterns}
\end{table}

\subsection{Minimal Gold Evidence Is Not Always the Best Context}

HippoRAG2 narrows the gap to oracle evidence in the single-document setting despite returning a much noisier pack. Oracle evidence contains the minimal annotated support, while HippoRAG2 often supplies additional local context. This suggests that a useful evidence pack for answer generation may differ from the smallest annotated sufficient set. Some non-gold context can help the answer model interpret quotations, resolve references, or connect local details to the query.

\subsection{Granularity Creates a Robustness--Detail Tradeoff}

The comparison between HippoRAG2 and LLM-Wiki evidence construction illustrates a granularity dilemma. Fine-grained graph retrieval is strong on local evidence and moderate aggregation, but performance declines as evidence must be collected from more documents. Coarser LLM-Wiki organization sacrifices single-document detail, but its thematic aggregation makes performance more stable as fan-in increases. This points toward hybrid evidence construction: coarse structures can identify the relevant thematic region, while fine-grained retrieval can extract the source passages needed for a faithful answer.

\subsection{Failure Case Studies}
\label{app:case-studies}

We present two contrastive cases that illustrate why HippoRAG2 and LLM-Wiki fail on complementary instance sets (Jaccard overlap of failure sets = 0.13). Each case shows the gold evidence and claims alongside the two systems' answers, making the failure mechanism directly observable.

\begin{figure*}[t]
\centering
\includegraphics[width=\textwidth]{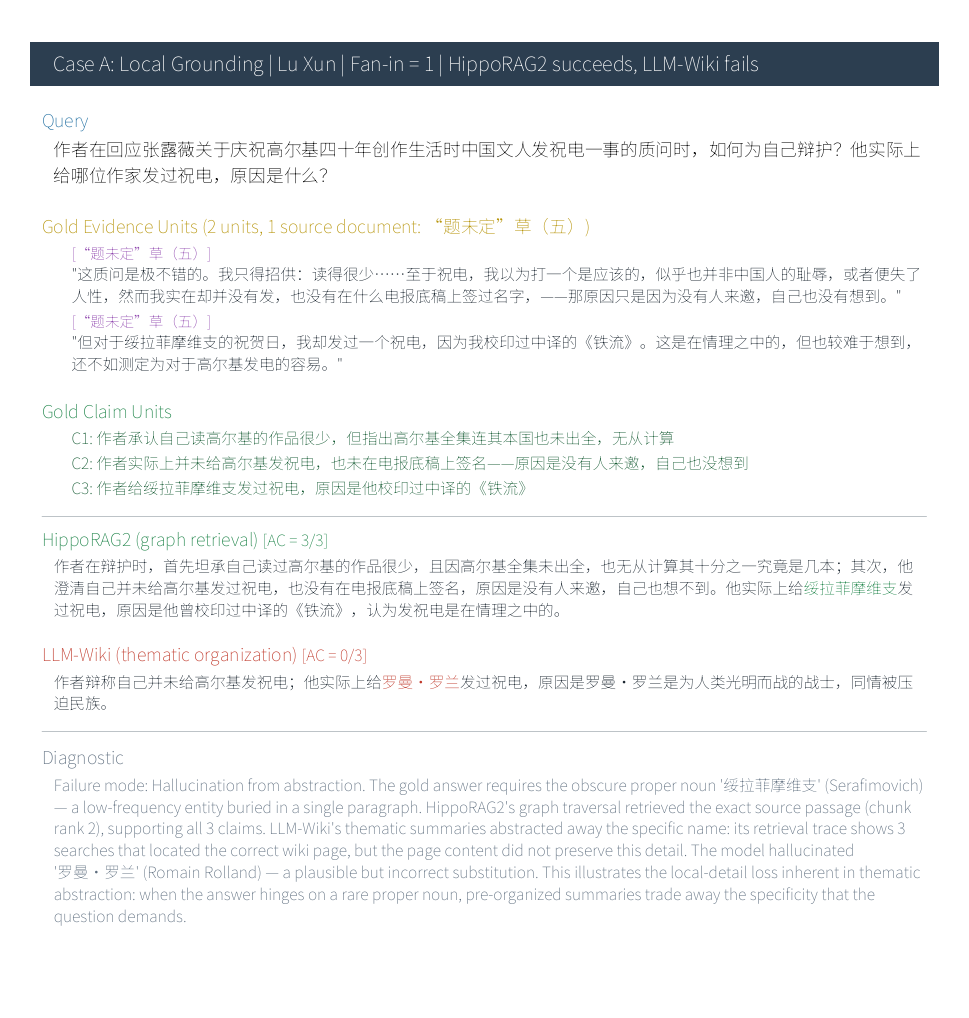}
\caption{Case A: HippoRAG2 succeeds, LLM-Wiki fails (Lu Xun, fan-in = 1). Graph retrieval locates the exact passage containing the rare entity; thematic organization abstracts it away and the model hallucinates a plausible but incorrect name. All Chinese text appears only in the embedded figure.}
\label{fig:case-single}
\end{figure*}

\paragraph{Case A (Figure~\ref{fig:case-single}): Local grounding favors graph retrieval.}
A single-document question about Lu Xun asks which author received a congratulatory telegram and why. The gold answer requires an obscure transliterated proper noun (Serafimovich) buried in one paragraph of a single essay. HippoRAG2's graph traversal retrieved the exact source passage at rank 2 and answered all three claims correctly (AC = 3/3). LLM-Wiki's thematic summaries had abstracted away this low-frequency entity: its retrieval trace shows three searches that located the correct wiki page title, but the page content did not preserve the specific name. The model hallucinated a different author name (Romain Rolland)---a plausible but incorrect substitution (AC = 0/3). This case exemplifies the \emph{local-detail loss} inherent in thematic abstraction: when the answer hinges on a rare proper noun within a single passage, pre-organized summaries trade away the specificity that the question demands.

\begin{figure*}[t]
\centering
\includegraphics[width=0.96\textwidth]{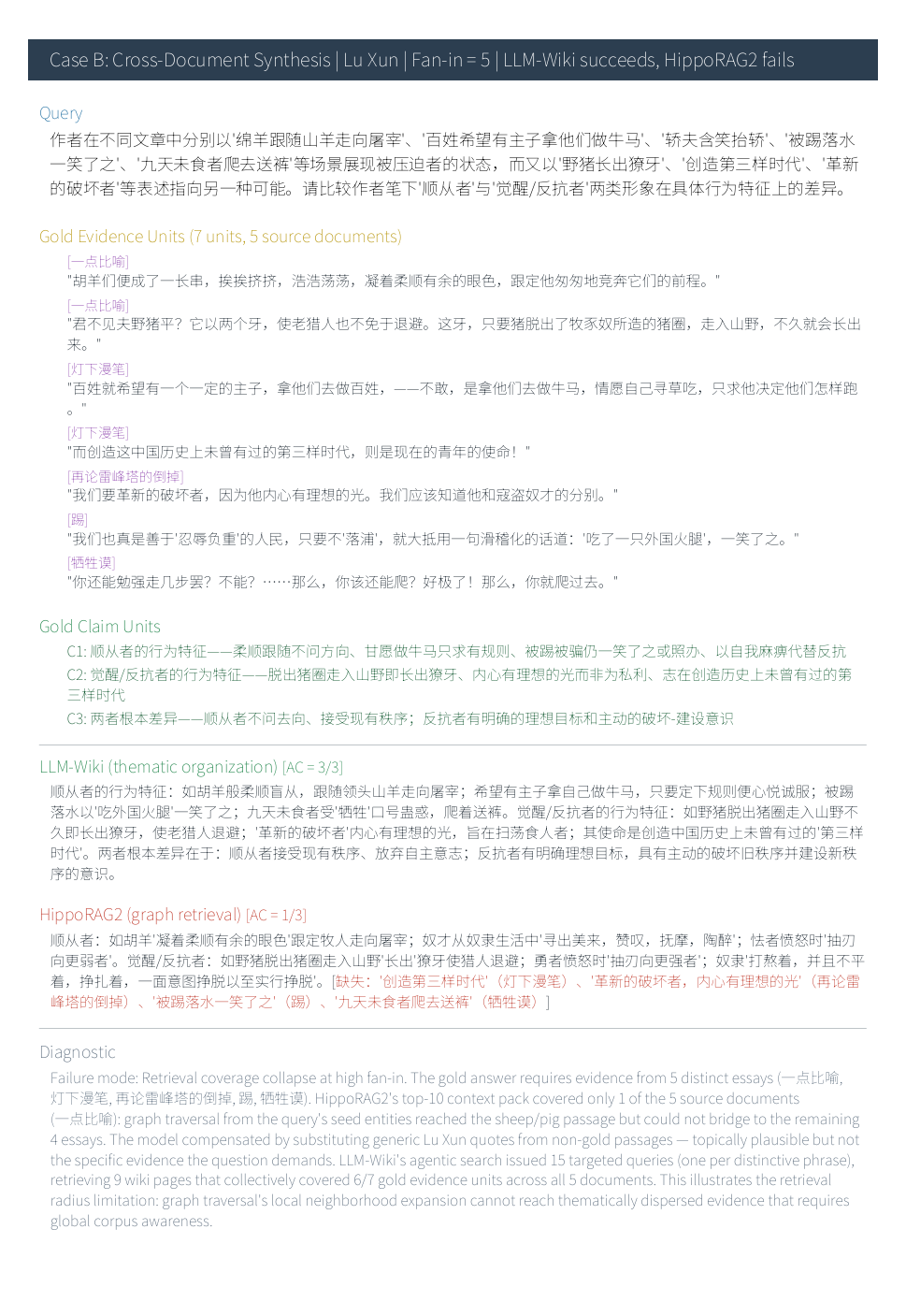}
\caption{Case B: LLM-Wiki succeeds, HippoRAG2 fails (Lu Xun, fan-in = 5). Thematic search covers all 5 source essays through targeted queries; graph traversal's context pack spans only 1/5 gold documents and the model fills gaps with non-gold substitutes. All Chinese text appears only in the embedded figure.}
\label{fig:case-multi}
\end{figure*}

\paragraph{Case B (Figure~\ref{fig:case-multi}): Cross-document synthesis favors thematic organization.}
A high-fan-in question (fan-in = 5) asks the reader to compare ``submissive'' and ``awakened'' archetypes across five Lu Xun essays. LLM-Wiki's agentic search issued 15 targeted queries---one per distinctive phrase in the question---and retrieved 9 wiki pages that collectively covered 6 out of 7 gold evidence units from all 5 source documents (AC = 3/3). HippoRAG2's top-10 context pack covered only 1 of the 5 gold source documents: graph traversal from the query's seed entities reached the sheep/pig passage in one essay but could not bridge to the remaining four. The model compensated by substituting topically plausible but non-gold quotes from other essays (AC = 1/3). This case exemplifies the \emph{retrieval radius limitation}: graph traversal's local neighborhood expansion cannot reach thematically dispersed evidence that requires global corpus awareness.

\paragraph{Diagnostic summary.}
These two cases together demonstrate the complementary error structure underlying the crossover in Table~\ref{tab:crossover}: graph retrieval preserves rare local details that abstraction erases, while thematic organization provides the global coverage that local traversal cannot reach. The failure patterns are non-overlapping---Case~A is trivial for HippoRAG2 but impossible for LLM-Wiki, and vice versa for Case~B---which supports the fan-in-based system selection principle proposed in \S\ref{app:implications}.

\section{Actionable Implications}
\label{app:implications}

\paragraph{Choose evidence organization by target fan-in.}
Single-document detail benefits from fine-grained retrieval and graph structure. Moderate cross-document questions align well with event- or relation-oriented memory organization. Broad synthesis benefits from thematic aggregation, even when such aggregation lowers local precision.

\paragraph{Prioritize recall under generous context budgets.}
When raw retrieval is recall-limited and the answer model has enough context capacity to tolerate some noise, retrieval should favor covering the required evidence over aggressively filtering for a small pack. Precision-oriented filtering is most valuable when it preserves recall or converts raw evidence into a more usable structured representation.

\paragraph{Treat precision as evidence usability, not merely token efficiency.}
The GPT-family ablation shows that high-precision structured packs can outperform higher-recall noisy packs. This makes EP a useful design signal for systems that summarize, cluster, or reorganize evidence before generation.

\paragraph{Use structure before generation.}
The weakness of fixed-order long-context prompting shows that evidence construction should be treated as a distinct stage. Raw text exposure leaves selection and organization to the reader; explicit evidence construction externalizes these operations into a controllable pipeline that can be measured through ER, EP, and fan-in degradation.

\paragraph{Adopt adaptive granularity.}
No single representation dominates across fan-in levels. A practical system should estimate query fan-in and route among fine-grained retrieval, graph traversal, event memory, and thematic pages. A promising design is two-stage construction: use coarse thematic or event structures to identify the relevant region of the corpus, then use fine-grained retrieval to extract source passages for the final evidence pack.

\section{Generation Constraints and Prompts}
\label{app:prompts}

The generation prompt enforces the following constraints.
\begin{itemize}[leftmargin=*]
    \item The query must not mention article titles, title hints, title abbreviations, or explicit source-range phrases such as ``these articles.''
    \item The query must refer to the writer as ``the author'' rather than using the author's name.
    \item Each multi-document question must genuinely require multiple documents; if one document is sufficient, the instance is invalid.
    \item Questions should be bounded by concrete cross-document details rather than broad prompts such as ``what is the author's overall view of X.''
    \item The answer must be directly supported by the evidence and should avoid unsupported literary interpretation beyond the quoted spans.
    \item \texttt{gold\_evidence\_units} must be direct quotations from the input documents and use only allowed document IDs.
    \item \texttt{gold\_claim\_units} must be atomic, checkable conclusions that the reference answer should cover.
\end{itemize}

\section{Instance Generation Prompts}
\label{app:generation-prompts}

This section provides the full generation prompt templates used for single-document and multi-document instance construction. All prompts are originally written in Chinese; we present English translations below for reviewer readability. The original Chinese versions are used in all generation runs. The generation model is Claude Opus 4.6 with temperature 0.2 and a maximum generation budget of 24,000 tokens.

\subsection{Single-Document Generation Prompt}
\label{app:single-doc-prompt}

Each single-document generation call receives one article and produces up to 3 instances grounded in that article alone. The system prompt establishes the annotation protocol and quality constraints; the user prompt provides the article text and metadata.

\promptheading{System prompt.}

\begin{promptbox}
\small
\texttt{You are an extremely strict Chinese literature benchmark annotator.}

\texttt{Your task: Given the full text of a single article, directly generate single-document instances that conform to the annotation schema.}

\texttt{\textbf{I. Task Nature}}

\texttt{This is a Focused Reading scenario for single-document evidence construction.}

\texttt{Each generated instance must satisfy:}\\
\texttt{1. It can be answered primarily from this article alone.}\\
\texttt{2. The question must be grounded in concrete textual details in the article.}\\
\texttt{3. The answer must be supported by direct evidence from this article.}\\
\texttt{4. The instance should be suitable as a single-document sample for an author-grounded QA benchmark.}

\texttt{\textbf{II. Query Requirements}}

\texttt{1. The query must NOT contain the article title, title abbreviations, title hints, or any information that would reveal which article to retrieve.}\\
\texttt{2. Always refer to the writer as ``the author''; never use the author's real name or pen name.}\\
\texttt{3. Do not use explicit range indicators such as ``in this article'' or ``in the text''; instead, implicitly delimit the scope through textual details.}\\
\texttt{4. A question may only be generated if it can be stably anchored by unique details in the current article.}\\
\texttt{5. If a question would likely hold equally well in many other articles by the same author, do not generate it.}\\
\texttt{6. By default, do not generate overly abstract questions with high generalization risk, such as:}\\
\texttt{~~~- Overly broad central-idea questions}\\
\texttt{~~~- Overly broad writing-motivation questions}\\
\texttt{~~~- Overly broad author-attitude summary questions}\\
\texttt{~~~- Overly broad concept-definition questions}\\
\texttt{~~~unless the question is strongly anchored by highly specific, low-ambiguity in-text details.}\\
\texttt{6.1. Each query must contain exactly one main question; do not ask multiple aspects simultaneously.}

\texttt{\textbf{III. Evidence and Answer Requirements}}

\texttt{7. gold\_evidence\_units must be direct quotations from the article; no paraphrasing, no summarizing.}\\
\texttt{8. Each element in gold\_evidence\_units must be: \{"doc\_id": "...", "text": "..."\}.}\\
\texttt{9. gold\_claim\_units must be written as ``core conclusion units the reference answer should cover''---clear, verifiable, and appropriately granular.}\\
\texttt{10. reference\_answer must be concise, definitive, and supportable by gold\_evidence\_units.}\\
\texttt{11. gold\_claim\_units should not merely paraphrase the reference\_answer; they should be the genuine core information points that the answer must cover.}\\
\texttt{12. query, gold\_claim\_units, and reference\_answer must be mutually consistent.}

\texttt{\textbf{IV. Output Format}}

\texttt{Output only the following JSON structure. No explanations, no Markdown, no code fences.}\\
\texttt{Top-level must be: \{"instances": [\{"query": "...", "gold\_evidence\_units": [...], "gold\_claim\_units": [...], "reference\_answer": "..."\}]\}}\\
\texttt{If the article cannot stably produce high-quality instances, output fewer; prefer quality over quantity.}
\end{promptbox}

\promptheading{User prompt template.}

\begin{promptbox}
\small
\texttt{Please generate single-document instances based on the following article.}

\texttt{Metadata (for your understanding only; do not repeat in output):}\\
\texttt{- author\_id: \{author\_id\}}\\
\texttt{- theme: \{theme\}}\\
\texttt{- Allowed doc\_id: \{doc\_id\}}\\
\texttt{- Article title (for internal reference only; must NOT appear in query): \{doc\_title\}}

\texttt{Notes:}\\
\texttt{1. This is single-document; all evidence must come from this article.}\\
\texttt{2. All doc\_id in gold\_evidence\_units must be: \{doc\_id\}.}\\
\texttt{3. Query must avoid title leakage and explicit range indicators.}\\
\texttt{4. Always use ``the author'' to refer to the writer.}\\
\texttt{5. gold\_evidence\_units must be direct quotations.}\\
\texttt{6. gold\_claim\_units must be core conclusion units.}\\
\texttt{7. reference\_answer must be a concise standard answer supportable by the evidence.}

\texttt{Suggested number of instances: up to \{instances\_per\_doc\}.}

\texttt{Full article text:}\\
\texttt{\{article\_text\}}

\texttt{Output only JSON.}
\end{promptbox}

\subsection{Multi-Document Generation Prompt}
\label{app:multi-doc-prompt}

Multi-document generation operates in two passes over same-theme document groups: a \emph{low-fan-in pass} (input: 5 articles, target evidence from 2--3 documents) and a \emph{high-fan-in pass} (input: 8 articles, target evidence from 5+ documents). Each pass produces up to 4--5 instances per group. The system prompt shares the annotation protocol with the single-document prompt but adds cross-document constraints and task-type requirements.

\promptheading{System prompt (shared structure; fan-in mode block varies).}

\begin{promptbox}
\small
\texttt{You are an extremely strict Chinese literature benchmark annotator.}

\texttt{Your task: Given the full text of multiple articles under the same theme, directly generate multi-document instances that conform to the annotation schema.}

\texttt{\textbf{I. Task Nature}}

\texttt{This is a Cross-Document Construction scenario.}

\texttt{Each instance must belong to one of three task types:}\\
\texttt{1. thematic\_synthesis --- Integrate information from multiple articles around the same theme to form an inductive answer.}\\
\texttt{2. contrastive\_reasoning --- Organize evidence around two or more objects, cases, or approaches to form a comparative structure.}\\
\texttt{3. diachronic\_evolution --- Only permitted when there exist describable stage differences, temporal ordering, and phase-level changes.}

\texttt{\textbf{II. Query Requirements}}

\texttt{1. Query must NOT contain any article titles, title abbreviations, or retrieval hints.}\\
\texttt{2. Always use ``the author''; never use the author's name.}\\
\texttt{3. Do not use explicit range indicators such as ``in these articles''; instead, implicitly delimit scope through combinations of textual details.}\\
\texttt{4. Each instance must genuinely require multiple articles for joint support; if any single article suffices, do not generate.}\\
\texttt{5. If a question merely looks like a synthesis question but can actually be answered from one article alone, do not generate.}\\
\texttt{6. By default, do not generate overly broad questions such as:}\\
\texttt{~~~- How does the author view a certain broad topic}\\
\texttt{~~~- What is the author's consistent attitude}\\
\texttt{~~~- What does the author advocate long-term}\\
\texttt{~~~unless the query is clearly narrowed by low-ambiguity cross-document details.}\\
\texttt{7. Queries should preferably ask about concrete phenomena, examples, statements, or judgments that can be directly located across multiple articles, rather than requiring the answerer to construct a grand interpretive framework.}\\
\texttt{8. The answer should be easily verifiable by different annotators after reading the evidence; if the answer requires extensive literary interpretation or value judgment, do not generate.}\\
\texttt{9. Details in the query should mainly serve to bound the question, not to pre-reveal the full answer structure.}\\
\texttt{10. Each query must contain exactly one main question.}\\
\texttt{11. Keep query length restrained; given the above requirements, the question should be as concise as possible.}

\texttt{\textbf{[Fan-in mode block]}}

\texttt{Low-fan-in mode: Prioritize generating instances that genuinely depend on 2--3 articles for joint support. If a question requires 4 or more articles, do not generate in this round.}

\texttt{High-fan-in mode: Prioritize generating instances that genuinely depend on 5--6 or more articles for joint support. If a question can be answered with fewer than 5 articles, do not generate in this round.}

\texttt{\textbf{III. Evidence and Answer Requirements}}

\texttt{12. gold\_evidence\_units must be direct quotations; no paraphrasing.}\\
\texttt{13. Each element: \{"doc\_id": "...", "text": "..."\}.}\\
\texttt{14. Evidence must be distributed across the multiple articles that the question actually requires; it must not concentrate in a single article.}\\
\texttt{15. gold\_claim\_units must be clear, verifiable, and reflect cross-document induction, comparison, or evolution structure.}\\
\texttt{16. reference\_answer must be concise, definitive, and supportable by the listed evidence.}\\
\texttt{17. reference\_answer may only perform within-evidence induction; do not add unsupported literary interpretation.}\\
\texttt{18. gold\_claim\_units should correspond to verifiable information points in the source text.}\\
\texttt{19. task\_type must be one of: thematic\_synthesis, contrastive\_reasoning, diachronic\_evolution.}\\
\texttt{20. thematic\_synthesis and contrastive\_reasoning should be roughly equal in quantity.}

\texttt{\textbf{IV. Output Format}}

\texttt{Output only JSON. Top-level: \{"instances": [\{"query": "...", "task\_type": "...", "gold\_evidence\_units": [...], "gold\_claim\_units": [...], "reference\_answer": "..."\}]\}}\\
\texttt{No explanations, no Markdown, no code fences. If the scope of the question is ambiguous, do not generate.}
\end{promptbox}

\promptheading{User prompt template.}

\begin{promptbox}
\small
\texttt{Please generate multi-document instances based on the following articles under the same theme.}

\texttt{Metadata (for your understanding only):}\\
\texttt{- author\_id: \{author\_id\}}\\
\texttt{- theme: \{theme\}}\\
\texttt{- Allowed doc\_ids (use ONLY these): \{allowed\_doc\_ids\}}

\texttt{\{fan\_in\_mode\_instruction\}}

\texttt{Notes:}\\
\texttt{1. All doc\_ids in gold\_evidence\_units must come from the allowed list above.}\\
\texttt{2. Each instance must genuinely require multiple articles for joint support.}\\
\texttt{3. task\_type must be one of: thematic\_synthesis, contrastive\_reasoning, diachronic\_evolution.}\\
\texttt{4. Query must not contain title information or explicit range indicators.}\\
\texttt{5. Always use ``the author'' to refer to the writer.}\\
\texttt{6. gold\_evidence\_units must be direct quotations.}\\
\texttt{7. gold\_claim\_units must be core conclusion units.}\\
\texttt{8. reference\_answer must be a concise standard answer supportable by the evidence.}\\
\texttt{9. Prefer questions with clear answer boundaries verifiable from the evidence.}\\
\texttt{10. reference\_answer should not over-elaborate; only summarize what the evidence stably supports.}

\texttt{Suggested number of instances: up to \{target\_instances\}.}

\texttt{Input articles:}\\
\texttt{[For each article: doc\_id, doc\_title (internal only), full text]}

\texttt{Output only JSON.}
\end{promptbox}

\subsection{Generation Configuration Summary}

Table~\ref{tab:gen-config} summarizes the generation hyperparameters.

\begin{table}[t]
\centering
\small
\begin{tabular}{ll}
\toprule
Parameter & Value \\
\midrule
Generation model & Claude Opus 4.6 \\
Temperature & 0.2 \\
Max generation tokens & 24,000 \\
Single-doc instances per article & up to 3 \\
Low-fan-in input articles per group & 5 \\
High-fan-in input articles per group & 8 \\
Low-fan-in instances per group & up to 4 \\
High-fan-in instances per group & up to 5 \\
Low-fan-in groups per theme & 24 \\
High-fan-in groups per theme & 12 \\
\bottomrule
\end{tabular}
\caption{Instance generation hyperparameters.}
\label{tab:gen-config}
\end{table}

\section{Evaluation Protocol and Judge Prompts}
\label{app:evaluation-protocol}

This section describes the full evaluation protocol used for Answer Correctness (AC), Evidence Recall (ER), and Evidence Precision (EP). All judge prompts are originally written in Chinese; we present English translations below for reviewer readability. The original Chinese versions are used in all reported experiments.

\subsection{Overview}

AuthTrace evaluates each system output through three LLM-based judges applied independently \citep{liu-etal-2023-g,NEURIPS2023_91f18a12}. For systems that produce evidence packs, predicted evidence is first normalized into canonical segments, then evaluated for ER (whether each gold evidence unit is covered) and EP (whether each predicted segment supports gold evidence). The predicted answer is evaluated separately for AC against gold claim units. Separating evidence evaluation from answer evaluation ensures that the diagnostic decomposition---whether failures originate from evidence construction or from answer synthesis---remains interpretable.

The judge model is Qwen3.5-397B-A17B in the main setting and GPT-4o-mini in the model-family ablation. All judge calls use temperature 0.0 for reproducibility. The maximum generation budget is 3,000 tokens for AC and 800 tokens for ER/EP.

\subsection{Evidence Normalization}
\label{app:evidence-normalization}

Predicted evidence packs vary in granularity across systems: flat RAG returns fixed-size chunks, graph retrieval may return full articles, LLM-Wiki systems return structured pages, and memory systems return compressed entries. To enable shared pack-level comparison, all predicted evidence is normalized into canonical segments before ER and EP evaluation.

The canonical segmentation procedure operates as follows:
\begin{enumerate}[leftmargin=*]
    \item Text is cleaned by normalizing encoding, collapsing whitespace, and standardizing quotation marks.
    \item The text is split into sentences at Chinese sentence-ending punctuation
    (\texttt{U+3002}, \texttt{U+FF01}, \texttt{U+FF1F}, \texttt{U+FF1B})
    and paragraph boundaries.
    \item Sentences exceeding 180 tokens are hard-split at the token level.
    \item Sentences are greedily merged into segments targeting approximately 120 tokens, with a maximum of 180 tokens and a minimum of 40 tokens.
    \item Trailing segments below the minimum threshold are merged with the preceding segment when the combined length remains under the maximum.
    \item Duplicate segments (by exact string match after normalization) are removed.
\end{enumerate}

Token counting uses a regex-based approximation: each Chinese character counts as one token, and each contiguous Latin alphanumeric string counts as one token. The maximum number of predicted evidence segments submitted to the judge per instance is capped at 2,000.

\subsection{Answer Correctness (AC) Judge}
\label{app:ac-prompt}

The AC judge uses a holistic 0--3 rubric that simultaneously evaluates gold claim coverage and irrelevant content. The judge receives the query, gold claim units, a reference answer, and the predicted answer, and outputs structured JSON containing per-claim judgments, an irrelevant-content assessment, and a final score with justification. The reported AC metric is the normalized score: $\text{AC} = \text{score} / 3 \times 100$.

\promptheading{System prompt.}

\begin{promptbox}
\small
\texttt{You are a strict benchmark answer judge.}

\texttt{You must complete two tasks:}\\
\texttt{1. Analyze the coverage of each gold\_claim\_unit (for diagnostics).}\\
\texttt{2. Assign a holistic 0--3 score based on the rubric (the final metric).}

\texttt{When scoring, you must consider two dimensions simultaneously:}\\
\texttt{- Dimension A: Coverage of gold claims.}\\
\texttt{- Dimension B: Whether the answer contains irrelevant, incorrect, or redundant content.}

\texttt{Output only JSON. Do not output anything else.}
\end{promptbox}

\promptheading{User prompt template.}

\begin{promptbox}
\small
\texttt{Please evaluate the quality of the Predicted Answer.}

\texttt{\textbf{Step 1: Analyze claim coverage item by item}}

\texttt{For each gold\_claim\_unit, determine whether the Predicted Answer covers it:}\\
\texttt{- supported: The claim is explicitly and correctly covered.}\\
\texttt{- partial: The claim is only partially covered or expressed too vaguely.}\\
\texttt{- missing: The claim is not covered.}\\
\texttt{- contradicted: The Predicted Answer clearly contradicts this claim.}

\texttt{\textbf{Step 2: Check for irrelevant content}}

\texttt{Check whether the Predicted Answer exhibits the following problems:}\\
\texttt{- Introduces topics or facts not asked by the question.}\\
\texttt{- Mixes in information from other articles or other people.}\\
\texttt{- Contains additional incorrect statements that contradict gold claims.}\\
\texttt{- Is lengthy and repetitive with large amounts of content unrelated to the core question.}

\texttt{\textbf{Step 3: Assign a holistic 0--3 score based on the rubric}}

\texttt{Scoring criteria (considering both claim coverage and irrelevant content):}

\texttt{3 (Excellent):}\\
\texttt{~~- All or nearly all core claims are correctly covered (supported).}\\
\texttt{~~- The answer is focused and accurate, with no significant irrelevant or incorrect content.}

\texttt{2 (Good):}\\
\texttt{~~- Most core claims are covered, with minor omissions or insufficient detail.}\\
\texttt{~~- Or: Claim coverage is fairly complete, but includes some irrelevant content}\\
\texttt{~~~~(not affecting core judgment).}

\texttt{1 (Partial):}\\
\texttt{~~- Only about half or fewer claims are covered, with clear omissions.}\\
\texttt{~~- Or: Contains substantial irrelevant/incorrect content that interferes with accuracy.}\\
\texttt{~~- Or: Mentions some relevant information, but key arguments are missing or off-target.}

\texttt{0 (Incorrect):}\\
\texttt{~~- Core claims are essentially uncovered.}\\
\texttt{~~- Or: Contains critical factual errors or direct contradictions with claims.}\\
\texttt{~~- Or: The answer is overwhelmed by irrelevant content; no useful information}\\
\texttt{~~~~can be extracted.}\\
\texttt{~~- Or: The answer is completely unrelated to the question.}

\texttt{Notes:}\\
\texttt{- Your primary basis for judgment is gold\_claim\_units.}\\
\texttt{- reference\_answer is only for auxiliary understanding; do not bypass claims}\\
\texttt{~~and score based on impression alone.}\\
\texttt{- Penalty for irrelevant content: A few extra background sentences (one or two}\\
\texttt{~~lines of setup) should not cause a deduction; but if irrelevant content}\\
\texttt{~~approaches or exceeds the volume of effective content, or contains factual}\\
\texttt{~~errors, the score should be downgraded.}

\texttt{[Query] \{query\}}

\texttt{[Gold Claim Units] \{gold\_claim\_units\_json\}}

\texttt{[Reference Answer] \{reference\_answer\}}

\texttt{[Predicted Answer] \{predicted\_answer\}}

\texttt{Please output:}\\
\texttt{\{}\\
\texttt{~~"claim\_judgments": [\{"claim": "...", "status": "supported|partial|}\\
\texttt{~~~~\texttt{missing|contradicted}", "reason": "one sentence"\}],}\\
\texttt{~~"irrelevant\_content": \{"has\_irrelevant": true/false,}\\
\texttt{~~~~"description": "brief description or 'none'"\},}\\
\texttt{~~"score": integer 0--3,}\\
\texttt{~~"reason": "one sentence combining claim coverage and irrelevant content"}\\
\texttt{\}}
\end{promptbox}

\subsection{Evidence Recall (ER) Judge}
\label{app:er-prompt}

The ER judge determines whether each gold evidence unit is semantically covered by the predicted context pack. For each gold evidence unit, the judge receives the query, the gold unit, and the full predicted context pack, then outputs a binary coverage decision. The coverage criterion is intentionally permissive with respect to surface form: longer excerpts, shorter excerpts, or paraphrased equivalents all count as coverage, as long as the key information of the gold evidence is preserved.

\promptheading{System prompt.}

\begin{promptbox}
\small
\texttt{You are a strict benchmark evidence judge.}

\texttt{Requirements:}\\
\texttt{1. Evidence Recall: Only assess whether gold evidence is covered by the}\\
\texttt{~~~predicted context.}\\
\texttt{2. Evidence Precision: Only assess whether predicted evidence matches}\\
\texttt{~~~gold evidence.}\\
\texttt{3. Do not conflate answer correctness with evidence quality.}\\
\texttt{4. Output only JSON.}
\end{promptbox}

\promptheading{User prompt template.}

\begin{promptbox}
\small
\texttt{Please determine: Is the given gold evidence unit covered by the predicted context?}

\texttt{Coverage criteria:}\\
\texttt{1. Verbatim identity is not required.}\\
\texttt{2. Longer excerpts, shorter excerpts, or essentially equivalent source}\\
\texttt{~~~passages are acceptable.}\\
\texttt{3. As long as the predicted context contains a passage sufficient to carry}\\
\texttt{~~~the key information of this gold evidence, judge as covered=1.}\\
\texttt{4. Otherwise covered=0.}

\texttt{[Query] \{query\}}

\texttt{[Gold Evidence Unit] \{gold\_evidence\_unit\_json\}}

\texttt{[Predicted Context Pack] \{predicted\_context\_pack\_json\}}

\texttt{Please output only JSON:}\\
\texttt{\{"covered": 0 or 1, "reason": "one sentence explanation"\}}
\end{promptbox}

Evidence Recall is computed as:
\begin{equation}
    \text{ER} = \frac{\sum_{i=1}^{|E^\star|} \texttt{covered}_i}{|E^\star|} \times 100\%.
\end{equation}

\subsection{Evidence Precision (EP) Judge}
\label{app:ep-prompt}

The EP judge determines whether each predicted evidence segment supports at least one gold evidence unit. For each canonical segment in the predicted pack, the judge receives the query, all gold evidence units, and the predicted segment, then outputs a binary match decision. The matching criterion allows predicted segments that contain extra context beyond the gold evidence to still count as matched, as long as the segment carries content that supports the query and gold evidence without being misleading.

\promptheading{User prompt template.}

The EP judge shares the same system prompt as the ER judge (Section~\ref{app:er-prompt}).

\begin{promptbox}
\small
\texttt{Please determine: Does this predicted evidence unit match gold evidence?}

\texttt{Matching criteria:}\\
\texttt{1. matched = 1: The predicted evidence unit can be aligned with, covers, or}\\
\texttt{~~~is essentially equivalent to at least one entry in gold\_evidence\_units.}\\
\texttt{2. matched = 0: The predicted evidence unit cannot be aligned with any}\\
\texttt{~~~gold evidence.}\\
\texttt{3. Do not automatically judge as 0 just because the predicted evidence unit}\\
\texttt{~~~contains extra context beyond the gold evidence; if it contains content}\\
\texttt{~~~that supports the query / gold evidence key information and the extra}\\
\texttt{~~~content is not clearly misleading, judge as matched = 1.}

\texttt{[Query] \{query\}}

\texttt{[Gold Evidence Units] \{gold\_evidence\_units\_json\}}

\texttt{[Predicted Evidence Unit] \{predicted\_evidence\_unit\_json\}}

\texttt{Please output only JSON:}\\
\texttt{\{"matched": 0 or 1, "matched\_gold\_indices": [indices from 0; empty if none],}\\
\texttt{~~"reason": "one sentence explanation"\}}
\end{promptbox}

Evidence Precision is computed as:
\begin{equation}
    \text{EP} = \frac{\sum_{j=1}^{|\hat{E}_q|} \texttt{matched}_j}{|\hat{E}_q|} \times 100\%.
\end{equation}

\subsection{Evaluation Configuration Summary}

Table~\ref{tab:eval-config} summarizes the key evaluation hyperparameters used in all experiments.

\begin{table}[t]
\centering
\small
\begin{tabular}{ll}
\toprule
Parameter & Value \\
\midrule
Judge model (main) & Qwen3.5-397B-A17B \\
Judge model (ablation) & GPT-4o-mini \\
Temperature & 0.0 \\
Max generation tokens (AC) & 3,000 \\
Max generation tokens (ER/EP) & 800 \\
Segment target length & $\sim$120 tokens \\
Segment maximum length & 180 tokens \\
Segment minimum length & 40 tokens \\
Max pred. segments per instance & 2,000 \\
\bottomrule
\end{tabular}
\caption{Evaluation protocol hyperparameters.}
\label{tab:eval-config}
\end{table}

\section{Evaluation Reliability}
\label{sec:evaluation-reliability}

We validate the LLM-based evaluation through two complementary analyses: inter-judge agreement between two LLM judges, and human--LLM agreement on a manually annotated subset.

\subsection{Inter-Judge Agreement}

To verify that diagnostic conclusions are stable across judge models, we evaluate the same system outputs with both Qwen3.5-397B-A17B and GPT-4o-mini on the 20\% stratified sample (419 instances). Pooled inter-judge Spearman $\rho = 0.74$ and 98.6\% within-$\pm$1 agreement confirm that system rankings are preserved despite an absolute calibration offset (GPT scores +10.5pp higher on average). This offset affects absolute AC values but preserves the relative ordering and fan-in degradation patterns that our diagnostic conclusions depend on.

\subsection{Human--LLM Agreement}

We further validate the LLM judge against human annotations. For the primary Qwen judge, one annotator independently scored 200 randomly sampled instances using the same 0--3 AC rubric. For the GPT judge, an additional 100 instances were annotated using the same protocol. For ER and EP, the annotator reviewed 1,000 randomly sampled per-unit judgments each (ER: sampled from 320 instances; EP: sampled from 250 instances), providing binary agree/disagree labels.

\begin{table}[t]
\centering\small
\setlength{\tabcolsep}{4pt}
\begin{tabular}{lccc}
\toprule
Metric & Spearman $\rho$ & Binary Agr.\ (\%) & $\kappa_w$ \\
\midrule
AC (human vs.\ Qwen) & 0.78 & 82.0 & 0.71 \\
AC (human vs.\ GPT) & 0.76 & 79.0 & 0.68 \\
ER (per-unit binary) & -- & 91.2 & -- \\
EP (per-unit binary) & -- & 88.4 & -- \\
\bottomrule
\end{tabular}
\caption{Human--LLM judge agreement. $\rho$: Spearman rank correlation on 0--3 AC scores. Binary Agr.: for AC, agreement on correct ($\geq$2) vs.\ incorrect ($\leq$1); for ER/EP, agreement on binary covered/matched decisions. $\kappa_w$: quadratic weighted Cohen's kappa on the 0--3 scale.}
\label{tab:human-agreement}
\end{table}

Table~\ref{tab:human-agreement} shows that human--LLM agreement ($\rho = 0.78$ for Qwen on 200 instances, $0.76$ for GPT on 100 instances) is comparable to or slightly exceeds inter-LLM agreement ($\rho = 0.74$). The binary agreement rates for ER (91.2\%) and EP (88.4\%) indicate that per-unit evidence judgments are reliable at the level needed for aggregate metric computation. The primary source of AC disagreement involves partial-credit instances (score 1 vs.\ 2), where boundary placement is inherently subjective; the binary correct/incorrect distinction is more robust (82\% agreement). Critically, disagreements do not systematically favor any particular system, confirming that the comparative conclusions---which system performs better at which fan-in level---are preserved under human evaluation.

\end{document}